\documentclass{article}

     \PassOptionsToPackage{numbers}{natbib}

     \usepackage[final]{neurips_2020}

\usepackage[utf8]{inputenc} 
\usepackage[T1]{fontenc}    
\usepackage{hyperref}       
\usepackage{url}            
\usepackage{booktabs}       
\usepackage{amsfonts}       
\usepackage{nicefrac}       
\usepackage{microtype}      

\usepackage{graphicx}
\usepackage{amsfonts}
\usepackage{amsmath}
\usepackage{framed}
\usepackage{color}
\usepackage{multirow}
\usepackage{subfigure}
\usepackage{caption}

\title{Interpreting and Disentangling Feature Components of Various Complexity from DNNs}
\author{
  Jie Ren {\small *}\\
  Shanghai Jiao Tong University\\
  \texttt{ariesrj@sjtu.edu.cn}\\
   \And
   Mingjie Li \thanks{Equal contribution.}\\
   Shanghai Jiao Tong University\\
   \texttt{LiMingjie0608@sjtu.edu.cn}\\
  \And
   Zexu Liu\\
   Shanghai Jiao Tong University\\
  \texttt{liuzexu@sjtu.edu.cn}\\
  \And
   Quanshi Zhang\\
   Shanghai Jiao Tong University\\
   \texttt{zqs1022@sjtu.edu.cn} \\
}

\begin{document}

\maketitle

\begin{abstract}
  This paper aims to define, quantify, and analyze the feature complexity that is learned by a DNN. We propose a generic definition for the feature complexity. Given the feature of a certain layer in the DNN, our method disentangles feature components of different complexity orders from the feature. We further design a set of metrics to evaluate the reliability, the effectiveness, and the significance of over-fitting of these feature components. Furthermore, we successfully discover a close relationship between the feature complexity and the performance of DNNs. As a generic mathematical tool, the feature complexity and the proposed metrics can also be used to analyze the success of network compression and knowledge distillation.
\end{abstract}

\section{Introduction}
\label{sec:introduction}
Deep neural networks (DNNs) have demonstrated significant success in various tasks. Besides the superior performance of DNNs, some attempts have been made to investigate the interpretability of DNNs in recent years. Previous studies of interpreting DNNs can be roughly summarized into two types, \emph{i.e.} the explanation of DNNs in a post-hoc manner~\cite{lundberg2017unified, ribeiro2016should}, and the analysis of the feature representation capacity of a DNN~\cite{higgins2017beta, achille2018emergence, achille2018information, fort2019stiffness, liang2019knowledge}.

This study focuses on a new perspective of analyzing the feature representation  capacity of DNNs. \emph{I.e.} we propose a number of generic metrics to define, quantify, and analyze the complexity of features in DNNs. Previous research usually analyzed the \textit{theoretical} maximum complexity of a DNN according to network architectures~\cite{arora2016understanding, zhang2016architectural, raghu2017expressive, boob2018complexity,manurangsi2018computational}. In comparison, we propose to quantify the \textit{real} complexity of features learned by a DNN, which is usually significantly different from the \textit{theoretical} maximum complexity that a DNN can achieve. For example, if we use a deep network to solve a linear regression problem, the theoretical complexity of features may be much higher than the real feature complexity. 

In this paper, for the feature of a specific intermediate layer, we define the real complexity of this feature as the minimum number of nonlinear transformations required to compute this feature. However, the quantification of nonlinear transformations presents significant challenges to state-of-the-art algorithms. Thus, we use the number of nonlinear layers to approximate the feature complexity. \emph{I.e.} if a feature component can be computed using $k$ nonlinear layers, but cannot be computed with $k-1$ nonlinear layers, we consider its complexity to be of the $k$-th order.

In this way, we can disentangle an intermediate-layer feature into feature components of different complexity orders, as Figure~\ref{fig:task} shows. The clear disentanglement of feature components enables the quantitative analysis of a DNN. We further investigate the reliability of feature components of different complexity orders, and explore the relationship between the feature complexity and the performance of DNNs. More specifically, we analyze DNNs from the following perspectives:

\begin{figure}[t]
		\begin{small}
		\begin{minipage}[t]{0.63\linewidth}
			\centering
			\includegraphics[width=0.95\linewidth]{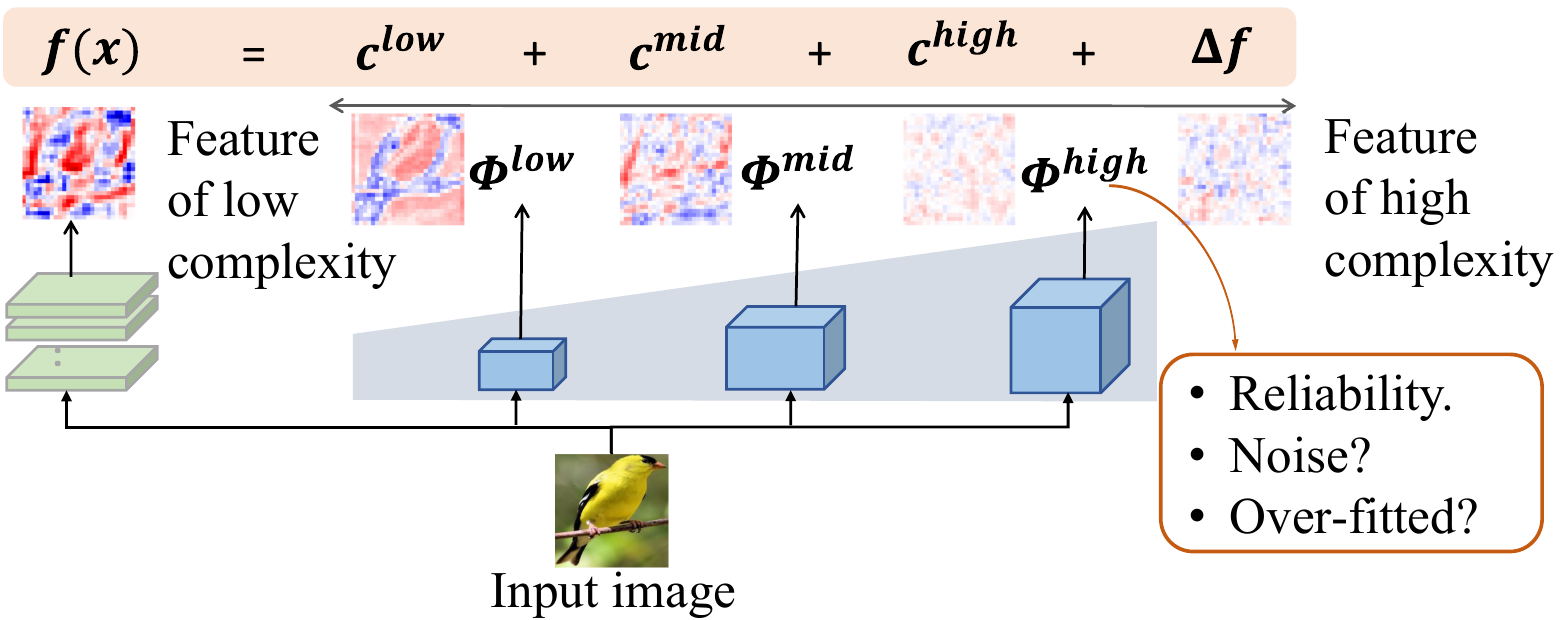}
			\caption{
				We disentangle the raw feature into feature components of different complexity orders. We further design metrics to analyze the disentangled feature components.
			}
			\label{fig:task}
		\end{minipage}
		\hfill
		\begin{minipage}[t]{0.35\linewidth}
			\centering
			\includegraphics[width=1.1\linewidth]{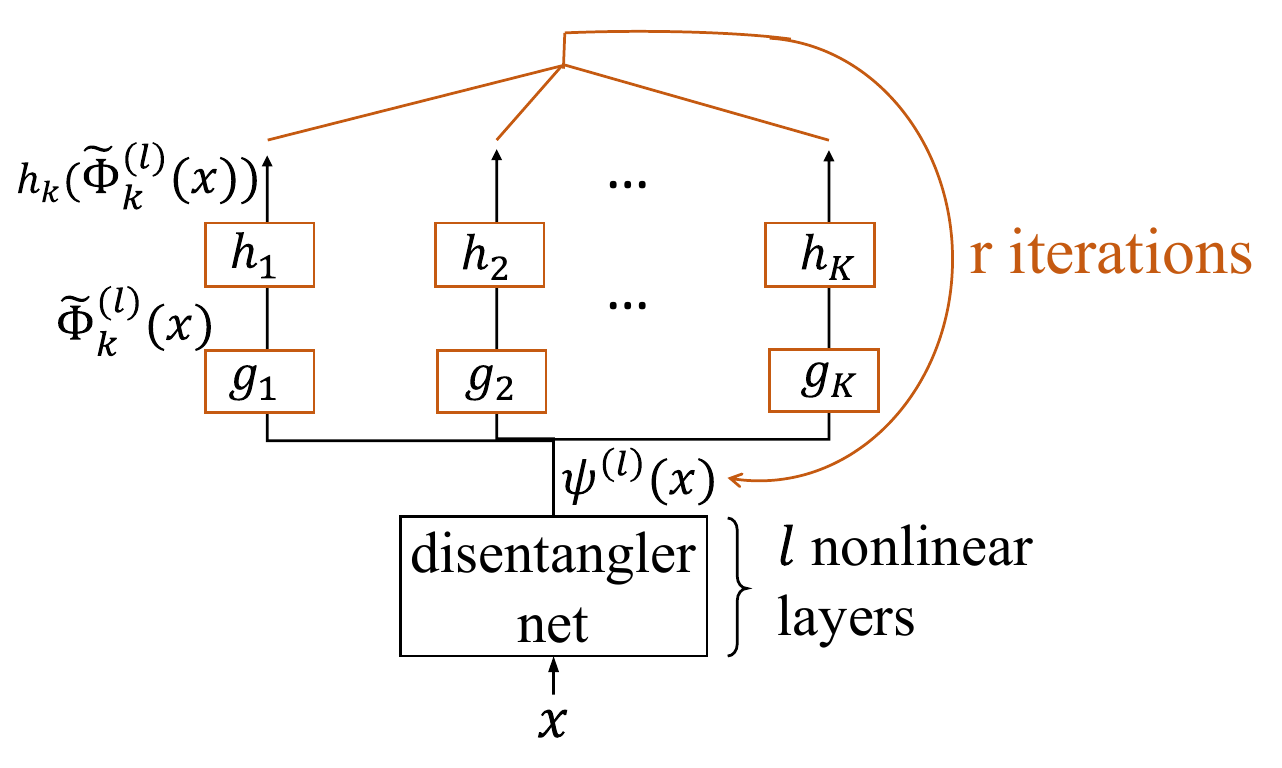}
			\vspace{-10pt}
			\caption{The network for the  disentanglement of reliable feature components.}
			\label{fig:reli-structure}
		\end{minipage}
	\end{small}
\end{figure}

\textbullet\quad The distribution of feature components of different complexity orders potentially reflects the difficulty of the task. 
A simple task usually makes the DNN mainly learn simple features.

\textbullet\quad We further define metrics to analyze the reliability, the effectiveness, and the significance of over-fitting for the disentangled feature components:
\begin{enumerate}
    \item In this paper, \textit{reliable} feature components refer to features that can be stably learned for the same task by DNNs with different architectures and parameters.
    
    \item \textit{The effectiveness of a feature component} is referred to as whether the feature component of a certain complexity order corresponds to neural activations relevant to the task. Usually, irrelevant neural activations can be considered as noises. 
    
    \item \textit{The significance of over-fitting of a feature component} represents that whether the feature component is over-fitted to specific training samples. In this paper, the significance of over-fitting is quantified as the difference between a feature component's numerical contribution to the decrease of the training loss and its contribution to the decrease of the testing loss.
    
    \item We successfully discover a strong connection between the feature complexity and the performance of DNNs.
\end{enumerate}

\textbullet\quad Taking the disentangled feature components  as the input feature of DNNs, especially feature components with high effectiveness and reliability, improves the performance of DNNs.

\textbf{Method:} More specifically, the disentanglement of feature components of different complexity orders is inspired by knowledge distillation~\cite{hinton2015distilling}. We consider the target DNN as the teacher network. Then, we design several disentangler networks (namely disentangler nets) with different depths to mimic the feature in an intermediate layer of the teacher network. Feature components mimicked by shallow disentangler nets usually correspond to those of low complexity.
A deeper disentangler net can incrementally learn an additional feature component of a bit higher complexity order, besides components of low complexity.

In addition, we find that the number of channels in disentangler nets does not significantly affect the distribution of feature components of different complexity orders. This demonstrates the trustworthiness of our method. The proposed method can be widely applied to DNNs learned for different tasks with different architectures. 
As generic mathematical tools, the proposed metrics provide insightful explanations for the success of network compression and knowledge distillation.

\textbf{Contributions:} Our contributions can be summarized as follows:
(1) We propose a method to define, quantify, and analyze the real complexity of intermediate-layer features in a DNN. Unlike the theoretical complexity of a DNN based on its architecture, the real feature complexity quantified in this paper reveals the difficulty of tasks. 
(2) The proposed method disentangles feature components of different complexity orders.
(3) We propose new metrics to analyze these feature components in terms of the reliability, the effectiveness, the significance of over-fitting, and the performance of DNNs. The analysis provides a new perspective to understand the network compression and the knowledge distillation.
(4) The disentangled feature components improve the performance of DNNs.

\section{Related Work}
In this section, we discuss related studies in the scope of interpreting DNNs.

\textbf{Visual explanations for DNNs:} The most direct way to interpret DNNs includes the visualization of the knowledge encoded in intermediate layers of DNNs~\cite{zeiler2014visualizing,simonyan2017deep,yosinski2015understanding,mahendran2015understanding, dosovitskiy2016inverting}, and the estimation of the pixel-wise attribution/importance/saliency on an input image~\cite{ribeiro2016should, lundberg2017unified, kindermans2017learning, fong2017interpretable, zhou2016learning, selvaraju2017grad, chattopadhay2018grad, zhou2014object}.
Some recent studies of network visualization reveal certain properties of DNNs. For example, \cite{fong2017interpretable} used influence functions to analyze the sensitivity to input data of a DNN. 

Unlike previous studies, in this paper, we propose to disentangle and visualize feature components of different complexity orders for better understandings of DNNs.

\textbf{Explanations for the representation capacity of DNNs:} The evaluation of the representation capacity of DNNs provides a new perspective for explanations. The information-bottleneck theory~\cite{wolchover2017new,shwartz2017opening} used mutual information to evaluate the representation capacity of DNNs~\cite{goldfeld2019estimating,xu2017information}. \cite{achille2018information} further used the information-bottleneck theory to constrain the feature representation during the learning process to learn more disentangled features. The CLEVER score~\cite{weng2018evaluating} was used to estimate the robustness of DNNs. 
The stiffiness~\cite{fort2019stiffness}, the Fourier analysis~\cite{xu2018understanding}, and the sensitivity metrics~\cite{novak2018sensitivity} were proposed and applied to analyze the generalization capacity of DNNs.
The canonical correlation analysis (CCA)~\cite{kornblith2019similarity} was used to measure the similarity between feature representations of DNNs. \cite{liang2019knowledge} investigated the knowledge consistency between different DNNs. \cite{chen2018learning} proposed instance-wise feature selection via mutual information for model interpretation.

Unlike previous methods, our research aims to explain a DNN from the perspective of feature complexity. In comparison, previous methods mainly analyzed the difficulty of optimizing a DNN,
~\cite{arora2016understanding,blum1989training,boob2018complexity}
the architectural complexity,
~\cite{zhang2016architectural}
and the representation complexity,
~\cite{liang2017fisher, cortes2017adanet,pascanu2013construct,raghu2017expressive}
which are introduced as follows:

\textbullet\quad Difficulty or computational complexity of optimizing a DNN: Some studies focus on the amount of computation, which is required to ensure a certain accuracy of a task. 
\cite{blum1989training,livni2014computational} proved that learning a neural network with one hidden layer was NP-hard in the realizable case.
\cite{arora2016understanding} showed that a ReLU network with a single hidden layer could be trained in polynomial time when the dimension of input was constant. \cite{boob2018complexity,manurangsi2018computational} proved that it was NP-hard to train a two-hidden layer feedforward ReLU neural network.
Based on topological concepts, \cite{bianchini2014complexity} proposed to evaluate the complexity of functions implemented by neural networks. \cite{rolnick2017power} focused on the number of neurons required to compute a given function for a network with fixed depth.

\textbullet\quad Complexity measures of the feature representation in DNNs:
\cite{pascanu2013construct, zhang2016architectural} proposed three architectural complexity measures for RNNs. \cite{raghu2017expressive}  proved the maximal complexity of features grew exponentially with depth. \cite{liang2017fisher, cortes2017adanet} measured the maximal complexity of DNNs with Rademacher complexity.

However, unlike the investigation of the theoretical maximal complexity of a DNN, we focus on the real complexity of the feature. We disentangle and visualize feature components of different complexity orders. In addition, we define and analyze the quality of the disentangled feature components, and successfully discover a strong connection between the feature complexity and the performance of DNNs.

\section{Algorithm}
\subsection{Complexity of feature components}
\label{sec:complexity}
Given an input image $x$, let {\small$f(x)\in \mathbb{R}^n$} denote the feature of a specific intermediate layer of the DNN.  {\small$y=g(f(x))\in \mathbb{R}^C$} is given as the output of the DNN. \emph{E.g.} $C$ denotes the number of categories in the classification task.  In this study, we define the complexity of feature components as the minimum number of nonlinear transformations that are required to compute the feature components. The disentanglement of feature components of different complexity orders in Figure~\ref{fig:task} can be represented as follows.
\begin{equation}
f(x) = c^{(1)}(x)+c^{(2)}(x)+\ldots+c^{(L)}(x) + \Delta f
\end{equation}
where {\small$c^{(l)}(x)$} denotes the feature component of the $l$-th complexity order (or, the \textit{$l$-order complexity} for short). {\small$\Delta f$} is the feature component with higher-order complexity.
\begin{framed}
	\textbf{Definition:} The feature component $c$ of the $l$-order complexity is defined as the feature component that can be computed using $l$ nonlinear layers, but cannot be computed with $l-1$ nonlinear layers.
	\emph{I.e.} {\small$l= \mathop{\arg\!\min}_{l'} \{\Phi^{(l')}(x)=c\}$}, where {\small$\Phi^{(l')}(\cdot)$} denotes a neural network with  $l'$ nonlinear transformation layers.
\end{framed}
\textit{Instead of directly disentangling the feature component $c^{(l)}$, we propose to use knowledge distillation to disentangle all feature components with the complexity of no higher than the $l$-th order, \emph{i.e.} {\small$\Phi^{(l)}(x) = \sum_{i=1}^{l}c^{(i)}(x)$}.} Given a trained DNN as the teacher network, we select an intermediate layer $f$ of the DNN as the target layer. {\small$\Phi^{(l)}(x) = \sum_{i=1}^{l}c^{(i)}(x)$} is disentangled using another DNN (termed the \textit{disentangler net}) with $l$ nonlinear layers.
The MSE loss {\small$\Vert f(x) - \Phi^{(l)}(x) \Vert ^2$} is used to force {\small$\Phi^{(l)}(x)$} to mimic the target feature {\small$f(x)$}, where {\small$f(x)$} denotes the feature of the teacher network. We use disentangler nets with different depths {\small$\Phi^{(1)}, \Phi^{(2)}, \ldots, \Phi^{(L)}$}, to disentangle feature components of different complexity orders. In this way, the feature component of the $l$-order complexity is given as:
	\begin{equation}
	Loss = \Vert f(x) - \Phi^{(l)}(x) \Vert ^2, \qquad c^{(l)}(x) = \Phi^{(l)}(x) - \Phi^{(l-1)}(x)
	\label{equ:l-depth}
	\end{equation}
In particular, {\small$c^{(1)}(x) = \Phi^{(1)}(x)$}.
Thus, {\small$f(x)$} is disentangled into two parts: {\small$f(x)=\Phi^{(L)}(x)+\Delta f$} where {\small$\Delta f$} denotes the feature component with a higher complexity order than $L$.

\textbf{Significance of feature components ({\small$\boldsymbol{\rho_c^{(l)}}$}):} Furthermore, we quantify the significance of feature components of different complexity orders as the relative variance of feature components. The metric is designed as {\small$\rho_c ^{(l)}  = Var[c^{(l)}(x)]/Var[f(x)]$}, where {\small$Var[c^{(l)}(x)] =  \mathbb{E}_x [\Vert c^{(l)}(x) - \mathbb{E}_{x'}	[c^{(l)}(x')]\Vert^2]$}.
For the fair comparison between different DNNs, we use the variance of {\small$f(x)$} to normalize {\small$Var[c^{(l)}(x)]$}. {\small$\rho_c^{(l)}$} represents the significance of the  $l$-th order complex feature component \emph{w.r.t.} the entire feature.

\textbf{Limitations: accurate estimation vs. fair comparison.}  Theoretically, if the teacher DNN has $D$ nonlinear transformation layers, the complexity of its features must be no higher than the $D$-th order, \emph{i.e.} {\small$\Phi^{(D')}(x) = f(x)$}, {\small$D'\le D$}. However, the optimization capacity for the learning of disentangler nets is limited. A disentangler net with $D$ nonlinear layers cannot learn all features encoded in {\small$f(x)$}. Thus, when {\small$\Phi^{(D')}\approx f(x)$} in real implementations, we have {\small$D'\ge D$}.

In this way, {\small$\rho_c^{(l)}$} measures the relative distribution of feature components of different complexity orders, instead of an accurate disentanglement of feature components.
Nevertheless, as Figure~\ref{fig:different_disentangler} shows, even if we use disentangler nets with different architectures, we still get similar distributions of feature components. This proves the trustworthiness of our method, and enables the fair comparison of feature complexity between different DNNs.

\textbf{Effectiveness of feature components ({\small$\boldsymbol{\alpha_{\textrm{effective}}^{(l)}}$})} measures whether the feature component {\small$c^{(l)}(x)$} extracted from the training sample $x$ directly contributes to the task.
The metric is defined based on the game theory. 
We first quantify the numerical contribution {\small$\varphi_l^\textrm{train}$} of each feature component {\small$c^{(l)}(x)$} to the decrease of the task loss in training as the Shapley value~\cite{shapley1953value,lundberg2017unified}. 
\emph{I.e.} $\varphi_l^\textrm{train}$ measures the change of the training loss caused by feature components {\small $\{c^{(l)}(x)|x\in X_\textrm{train}\}$}.
The Shapley value is an unbiased method to compute contributions of input features to the prediction, which satisfies four desirable axioms, \emph{i.e.}  efficiency, symmetry, linearity, and dummy axioms~\cite{grabisch1999axiomatic}. Please see the supplementary material for theoretical foundations of Shapley values. In this way, numerical contributions of all the {\small$L$} feature components can be allocated and given as {\small$\varphi_1^\textrm{train}+\varphi_2^\textrm{train}+\dots+\varphi_L^\textrm{train}=\mathbb{E}_{x\in X_\textrm{train}}[\mathcal{L}(\Delta f_x)-\mathcal{L}(\Delta f_x + \Phi^{(L)}(x))]$}, where {\small$\Delta f_x$} is the high-order component computed using the sample $x$. 
{\small$\mathcal{L}(\Delta f_x)$} represents the task loss when we remove all feature components in {\small$\Phi^{(L)}(x)$}, and {\small$\mathcal{L}(\Delta f_x+\Phi^{(L)}(x))$} denotes the task loss when both {\small$\Delta f_x$} and feature components in {\small$\Phi^{(L)}(x)$} are used for inference. Thus, the metric {\small$\alpha^{(l)}_{\mathrm{effective}}=\varphi_l^\textrm{train}/\sqrt{Var[c^{(l)}(x)]}$} measures the effectiveness of the feature component {\small$c^{(l)}$} to the decrease of the training loss. We use {\small$\sqrt{Var[c^{(l)}(x)]}$} for normalization.
Please see the supplementary material for theoretical foundations of the trustworthiness of {\small$\alpha^{(l)}_\textrm{effective}$}.

\textbf{Significance of over-fitting of feature components ({\small$\boldsymbol{\alpha_{\textrm{overfit}}^{(l)}}$})} measures whether {\small$c^{(l)}(x)$} is over-fitted to specific training samples. 
Similarly, this metric is also defined based on Shapley values. 
We quantify the numerical contribution {\small$\varphi_l^\textrm{overfit}$} of each feature component {\small$c^{(l)}(x)$} to over-fitting, whose significance is quantified as
{\small $\mathcal{L}_\textrm{overfit}(f)=\mathcal{L}_\textrm{overfit}(\Delta f+\Phi^{(L)})=\mathbb{E}_{x\in X_\textrm{test}}[\mathcal{L}(\Delta f_x+\Phi^{(L)}(x))]-\mathbb{E}_{x\in X_\textrm{train}}[\mathcal{L}(\Delta f_x + \Phi^{(L)}(x))]$}.
In this way, the numerical contribution can also be measured as Shapley values  {\small$\varphi_1^\textrm{overfit}+\varphi_2^\textrm{overfit}+\dots+\varphi_L^\textrm{overfit}=\mathcal{L}_\textrm{overfit}(\Delta f+\Phi^{(L)})-\mathcal{L}_\textrm{overfit}(\Delta f)$}, where {\small$\mathcal{L}_\textrm{overfit}(\Delta f + \Phi^{(L)})$} is computed using both components {\small $\Delta f_x$} and components {\small$\Phi^{(L)}(x)$} in different images.
\emph{I.e.} {\small$\varphi_l^\textrm{overfit}$} measures the change of {\small$\mathcal{L}_\textrm{overfit}$} caused by the feature component {\small$c^{(l)}(x)$}. The metric of the significance of over-fitting for {\small$c^{(l)}$} is given as {\small$\alpha^{(l)}_{\mathrm{overfit}}=\varphi_l^\textrm{overfit}/\varphi^\textrm{train}_l$}. 
Thus, {\small$\alpha_\textrm{overfit}^{(l)}$} represents the ratio of the increase of the gap $\Delta \mathcal{L}_\textrm{overfit}$ to the decrease of the training loss $\Delta \mathcal{L}_\textrm{train}$. Please see the supplementary material for theoretical foundations of the trustworthiness of {\small$\alpha^{(l)}_\textrm{overfit}$}.

\subsection{Reliability of feature components}
\label{sec:reliability}
In order to evaluate the reliability of a set of feature components {\small$\Phi^{(l)}(x) = \sum_{i=1}^{l}c^{(i)}(x)$}, we propose to disentangle reliable feature components {\small$\Phi^{(l),\textrm{reli}}(x)$} and unreliable feature components {\small$\Phi^{(l),\textrm{unreli}}(x)$}:
\begin{equation}
\Phi^{(l)}(x) = \Phi^{(l),\textrm{reli}}(x) + \Phi^{(l),\textrm{unreli}}(x)
\end{equation}

As discussed in~\cite{liang2019knowledge}, DNNs with different initializations of parameters usually learn some similar feature representations for the same task, and these similar features are proved to be reliable for the task.
Thus, we consider the reliable feature components as features that can be stably learned by different DNNs trained for the same task. Suppose that we have $K$ different DNNs learned for the same task. For each DNN, we select the feature of a specific intermediate layer as the target feature.
Let {\small$f_1(x), f_2(x), \ldots, f_K(x)$} denote target features of $K$ DNNs. We aim to extract features shared by {\small$f_1(x), f_2(x), \ldots, f_K(x)$}, \emph{i.e.} disentangling {\small$\Phi_1^{(l),\textrm{reli}}(x), \Phi_2^{(l),\textrm{reli}}(x), \ldots, \Phi_K^{(l),\textrm{reli}}(x)$} from features of $K$ DNNs as reliable components, respectively.
For each pair of DNNs {\small$(i,j)$}, {\small$\Phi_i^{(l),\textrm{reli}}(x)$} and {\small$\Phi_j^{(l),\textrm{reli}}(x)$} are supposed to be able to reconstruct each other by a linear transformation: 
\begin{equation}
\Phi_i^{(l),\textrm{reli}}(x) =r_{j \rightarrow i}(\Phi_j^{(l),\textrm{reli}}(x)),\quad \Phi_j^{(l),\textrm{reli}}(x) = r_{i \rightarrow j}(\Phi_i^{(l),\textrm{reli}}(x))
\end{equation}
where {\small$r_{i\rightarrow j}$} and {\small$r_{j \rightarrow i}$} denote two linear transformations.

\textbf{Implementations:} Inspired by the CycleGAN~\cite{zhu2017unpaired}, we apply the idea of cycle consistency on knowledge distillation to extract reliable feature components.
To extract reliable feature components, we construct the following neural network for knowledge distillation. As Figure~\ref{fig:reli-structure} shows, the network has a total of $l$ ReLU layers. We add $K$ parallel additional convolutional layers {\small$g_1,g_2,\ldots,g_{K}$} to generate $K$ outputs {\small$\widetilde{\Phi}^{(l)}_1(x), \widetilde{\Phi}^{(l)}_2(x),\ldots,\widetilde{\Phi}^{(l)}_{K}(x)$}, to mimic {\small$f_1(x), f_2(x),\ldots,f_{K}(x)$}, respectively. More specifically, {\small$\widetilde{\Phi}^{(l)}_k(x) = g_k(\psi^{(l)}(x))$}, where {\small$\psi^{(l)}(x)$} denotes the output of the dsentangler net with $l$ ReLU layers. Then, the distillation loss is given as {\small$\mathcal{L}^{\textrm{distill}} =\sum_{k=1}^{K} \Vert f_k(x)-\widetilde{\Phi}^{(l)}_k(x) \Vert^2$}.

For the cycle consistency, we use {\small$\widetilde{\Phi}^{(l)}_k(x)$} to reconstruct {\small$\psi^{(l)}(x)$} by another linear transformation {\small$h_k$}: {\small$h_k(\widetilde{\Phi}^{(l)}_k(x))=h_k(g_k(\psi^{(l)}(x)))\rightarrow\psi^{(l)}(x)$}.  We conduct cycle reconstructions between {\small$\psi^{(l)}(x)$} and {\small$\widetilde{\Phi}^{(l)}_k(x)$} for $R$ iterations ($R=10$ in experiments). Let {\small$\psi^{(l)}_{0}(x)= \psi^{(l)}(x), \psi^{(l)}_r(x)=\mathbb{E}_k[h_k\circ g_k\circ \psi^{(l)}_{r-1}(x)]$} denote the reconstruction output in the $r$-th iteration, where {\small$h_k\circ g_k$} denotes the cascaded layerwise operations. The cycle construction loss is given as follows:
\begin{equation}
\mathcal{L}^{\textrm{cycle}}= {\sum}_{r=1}^{R}{\sum}_{k=1}^{K}\Vert h_k\circ g_k \circ \psi^{(l)}_{r-1}(x)- \psi^{(l)}_{r-1}(x)\Vert ^2
\label{equ:loss of reli}
\end{equation}
Please see the supplementary material for the detailed explanation.

This loss makes the feature {\small$\widetilde{\Phi}^{(l)}_k(x)$} approximately shared by $K$ DNNs. In this way, {\small$\Phi_k^{(l),\textrm{reli}}(x)=\widetilde{\Phi}^{(l)}_k(x)$} can be considered as the reliable feature component. Compared with the traditional cycle consistency~\cite{zhu2017unpaired}, the above loss is much simpler and requires less computational cost. In this way, we can disentangle the unreliable feature component as {\small$\Phi^{(l),\textrm{unreli}}_{k}(x)=\Phi_{k}^{(l)}(x)-\Phi^{(l),\textrm{reli}}_{k}(x)$}. In experiments, in order to disentangle reliable and unreliable feature components from a target DNN, we used two additional trained DNNs $(A,B)$ to extract reliable feature components shared by the three DNNs, \emph{i.e.} $K=3$. DNNs $A$ and $B$ (namely \textit{exemplary DNNs}) were selected as those with state-of-the-art performance in the target task, in order to obtain convincing results. The same pair of DNNs $A$ and $B$ were uniformly used to analyze various DNNs, which enabled fair comparisons.

\textbf{Reliability of feature components}  
in $\Phi_k^{(l)}(x)$ can be quantified as the ratio of reliable feature components in  $\Phi_k^{(l)}(x)$ as  {\small$\rho^{(l),\textrm{reli}} = Var[\Phi_k^{(l),\textrm{reli}}(x)]/Var[\Phi^{(l)}_k(x)]$}.

\section{Experiments}

\textbf{Datasets, DNNs \& Implementation details:} We used our method to analyze VGG-16~\cite{simonyan2017deep} and ResNet-8/14/18/20/32/34/44~\cite{he2016deep}.\footnote{Compared with the original VGG-16, we added a BatchNorm layer before the output feature of each convolutional layer, before we use its feature to guide the distillation process. ResNet-8 and ResNet-14 had the similar structure as ResNet-20, ResNet-32 and ResNet-44 in ~\cite{he2016deep}, except that they had 1 and 2 blocks in each stage, respectively.} For simplification, we limited our attention to coarse-grained and fine-grained object classification. We trained these DNNs based on the CIFAR-10 dataset~\cite{krizhevsky2009learning}, the CUB200-2011 dataset~\cite{wah2011caltech}, and the Stanford Dogs dataset~\cite{KhoslaYaoJayadevaprakashFeiFei_FGVC2011}. 
For the CUB200-2011 dataset and the Stanford Dogs dataset, we used object images cropped by object bounding boxes for both training and testing. 
The classification accuracy of learned DNNs  is shown in the supplementary material.

\begin{figure}[t]
	\begin{small}
	\begin{minipage}[t]{0.47\linewidth}
		\centering
		\includegraphics[width=\linewidth]{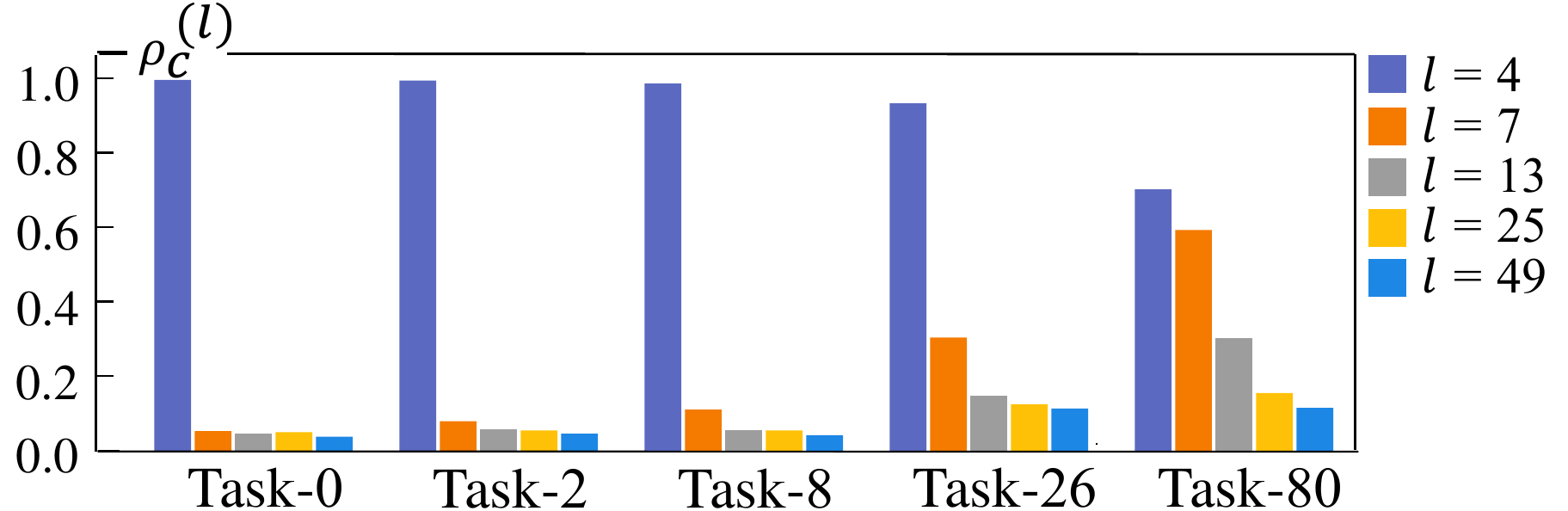}
		\vspace{-10pt}
		\caption{Significance of feature components of DNNs learned for different tasks.}
		\vspace{-13pt}
		\label{fig:task-feature}
	\end{minipage}	
	\hfill
	\begin{minipage}[t]{0.47\linewidth}
		\centering
		\includegraphics[width=0.95\linewidth]{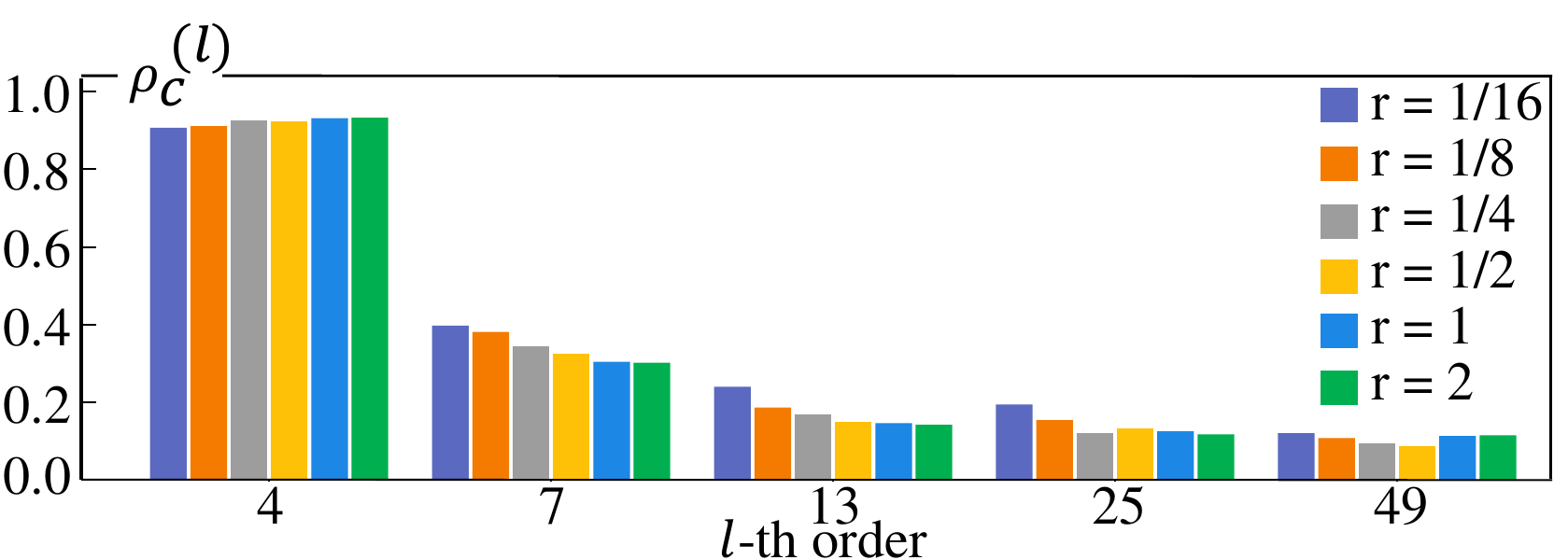}
		\caption{Significance of feature components disentangled by different disentangler nets.}
		\vspace{-13pt}
		\label{fig:different_disentangler}
	\end{minipage}
\end{small}
\end{figure}

\textbf{Disentangler nets:} We designed the disentangler nets {\small$\Phi^{(1)}(x),\dots,\Phi^{(L)}(x)$} with residual architectures. The disentangler net consisted of three types of residual blocks, each type having $m$ blocks. Each block of the three types consisted of a ReLU layer and a convolutional layer with {\small$128r, 256r ,512r$} channels,  respectively. In most experiments, we set {\small$r=1$}, but in Figure~\ref{fig:different_disentangler}, we tried different values of $r$ to test the performance of different disentangler nets. 
We used two additional convolutional layers before and after all $3m$ blocks, respectively, to match the input and output dimensions. Therefore, a disentangler net contained {\small$3m+2$} convolutional layers and {\small$l=3m+1$} ReLU layers.

For fair comparisons between DNNs, we used the same set of disentangler nets to quantify the complexity of each DNN. We analyzed the complexity of the output feature of the last convolutional layer. We set $m=1,2,4,8,16,32$, so that the non-linear layer numbers of disentangler nets were $l=4,7,13,25,49,97$. Considering the computational cost, we calculated {\small$c^{(4)}(x)\! =\! \Phi^{(4)}(x), c^{(7)}(x)\! =\! \Phi^{(7)}(x)\! -\! \Phi^{(4)}(x), c^{(13)}(x)\! =\! \Phi^{(13)}(x)\! -\! \Phi^{(7)}(x)$}, etc. 
This approximation did not affect the objectiveness of the quantified distribution of feature components of different complexity orders.

\begin{figure}[t]
	\begin{small}
	\begin{minipage}[t]{0.5\linewidth}
		\centering
		\includegraphics[width=\linewidth]{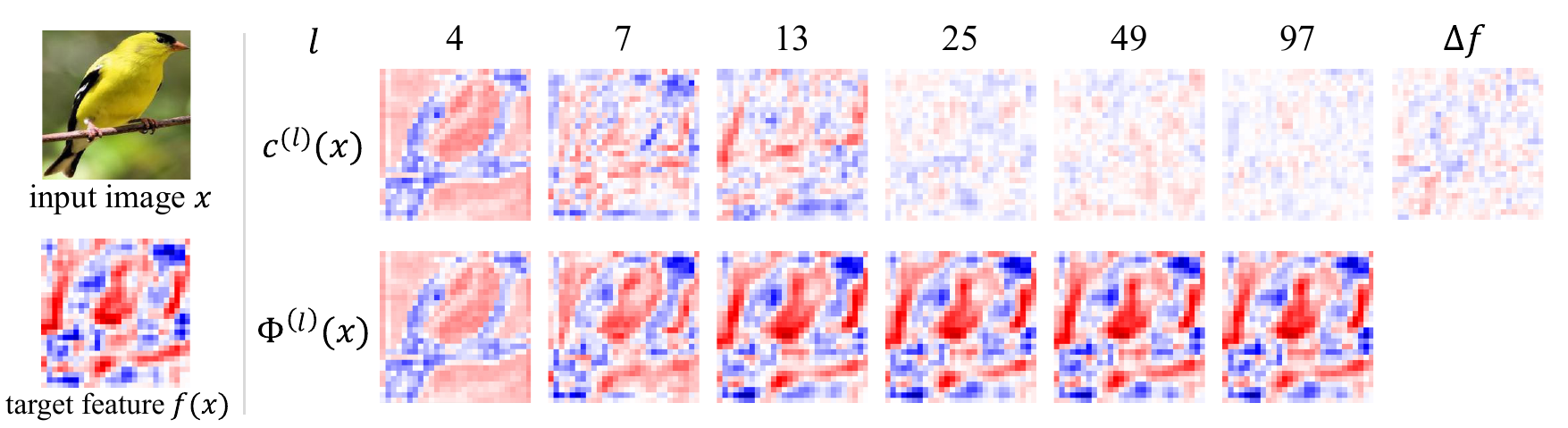}
	\end{minipage}
	\begin{minipage}[t]{0.5\linewidth}
		\centering
		\includegraphics[width=\linewidth]{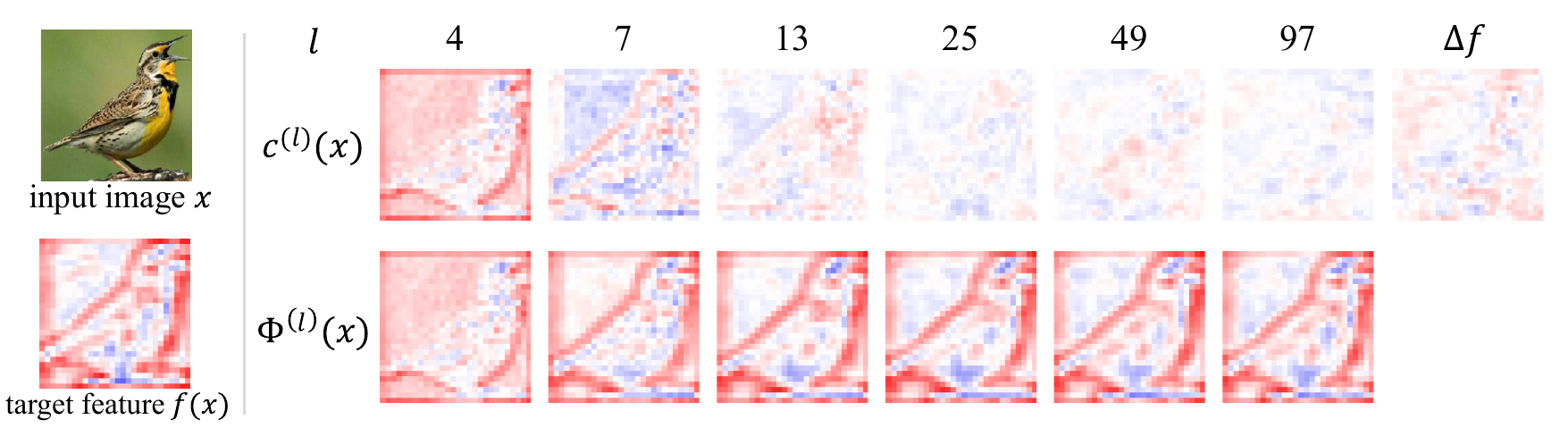}
	\end{minipage}
	\vspace{-15pt}
	\caption{Visualization of the disentangled feature components.}
	\vspace{-12pt}
	\label{fig:visualization of c}
\end{small}
\end{figure}

We visualized the disentangled feature components of different orders in Figure~\ref{fig:visualization of c}. Simple feature components usually represented general shape of objects, while complex feature components corresponded to detailed shape and noises.

\textbf{Exp. 1, task complexity vs. feature complexity: DNNs learned for simpler tasks usually encoded more feature components of low complexity orders.}
We defined tasks of different complexity orders . Let \textit{Task-$n$} denote a task of the $n$-order complexity as follows: we constructed another network (namely the task DNN) with $n$ ReLU layers and randomly initialized parameters, whose output was an {\small$8\times8\times64$} tensor. We learned the target DNN\footnote{For simplicity, we designed the target DNN to have the same architecture as the disentangler net with $l=19$.} to reconstruct this output tensor via an MSE loss. Since the task DNN contained $n$ ReLU layers, we used Task-$n$ to indicate the complexity of mimicking the task DNN.

Figure~\ref{fig:task-feature} compares distributions of feature components disentangled from target DNNs learned for Task-0, Task-2, Task-8, Task-26, and Task-80, respectively. DNNs learned for more complex tasks usually encoded more complex feature components.

\textbf{Various disentangler nets generated similar distributions of feature components, which demonstrated the trustworthiness of our method.} We learned a target DNN for Task-26 on the CIFAR-10 dataset and disentangled feature components from the output feature of the target DNN. We used disentangler nets with different architectures (different values of $r$) for analysis.
Figure~\ref{fig:different_disentangler} compares the distribution of feature components disentangled by different disentangler nets.

\textbf{Exp. 2, the number of training samples had small influence on the distribution of feature components, but significant impacts on the feature reliability.}
We learned ResNet-8/14/20/32/44 using different numbers of training samples, which were randomly sampled from the the CIFAR-10 dataset. Then, we disentangled feature components of different complexity orders from the output feature of the last residual block. More specifically, two exemplary DNNs $A$ and $B$ were implemented as ResNet-44 learned on the entire CIFAR-10 dataset with different initial parameters.

\begin{figure}[t]
	\begin{small}
		\subfigure{\includegraphics[width=0.48\linewidth]{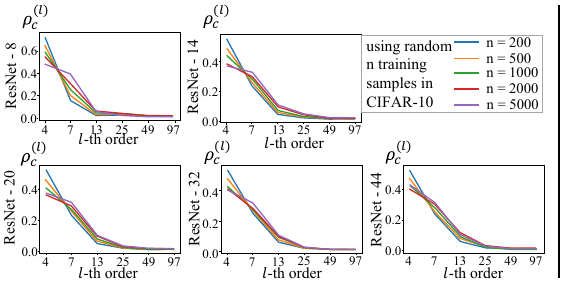}}
		\subfigure{\includegraphics[width=0.47\linewidth]{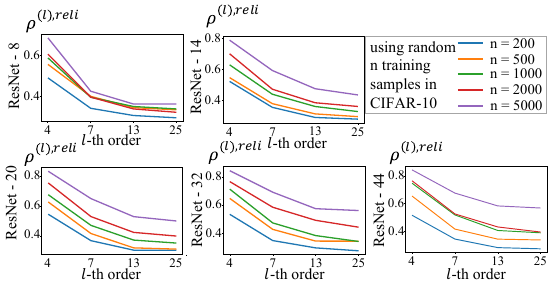}}
		\vspace{-10pt}
		\caption{Significance ({\small$\rho_c^{(l)}$}) and reliability ({\small$\rho^{(l),\textrm{reli}}$}) of the disentangled feature components.}
		\label{fig:analysis of c}
		\vspace{-10pt}
	\end{small}
\end{figure}

Figure~\ref{fig:analysis of c} compares the signficance of disentangled feature components {\small$\rho_c^{(l)}$} and the reliability of feature components {\small$\rho^{(l),\textrm{reli}}$} in different DNNs.
The DNN learned from the larger training set usually encoded more complex features, but the overall distribution of feature components was very close to the DNN learned from the smaller training set. This indicated that the number of training samples had small impacts on the significance of feature components of different complexity orders. 
However, in Figure~\ref{fig:analysis of c} (right), DNNs learned from many training samples always exhibited higher reliability than DNNs learned form a few training samples, which meant that the increase of the number of training samples would help DNN learn more reliable features. Results on the CUB200-2011 dataset and the Stanford Dogs dataset are shown in the supplementary material.

\begin{figure}
		\begin{small}
		\centering
		\subfigure{\includegraphics[height=.24\linewidth]{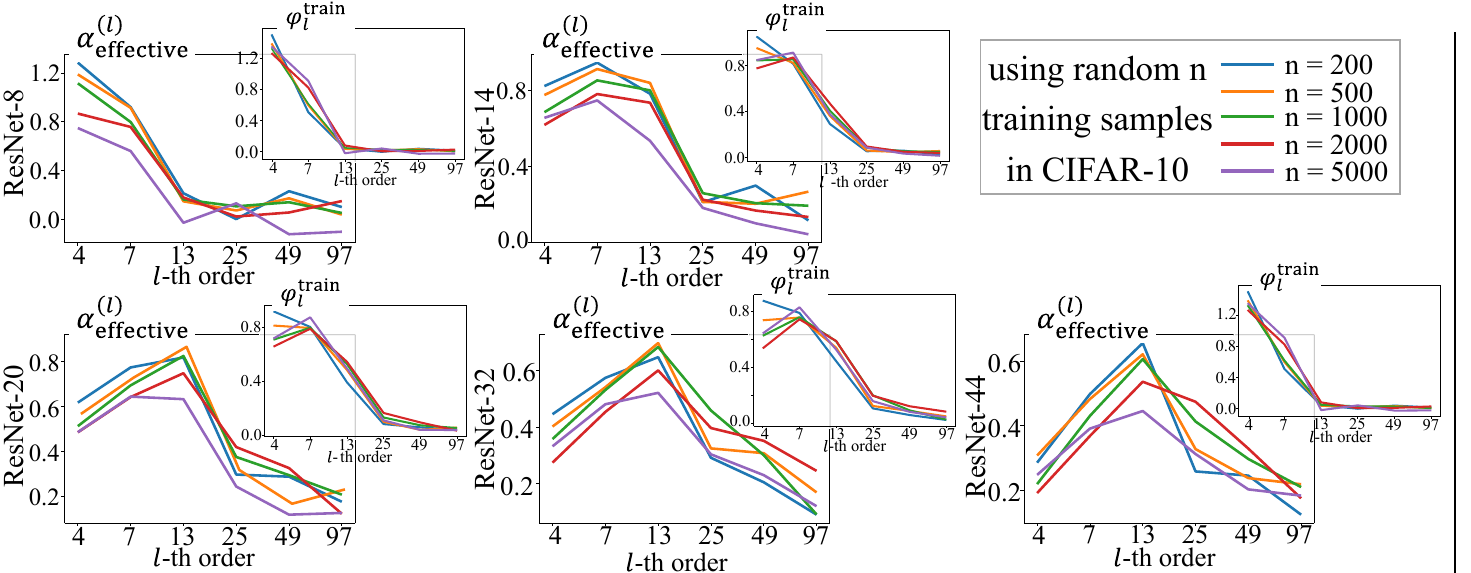}}
		\subfigure{\includegraphics[height=.24\linewidth]{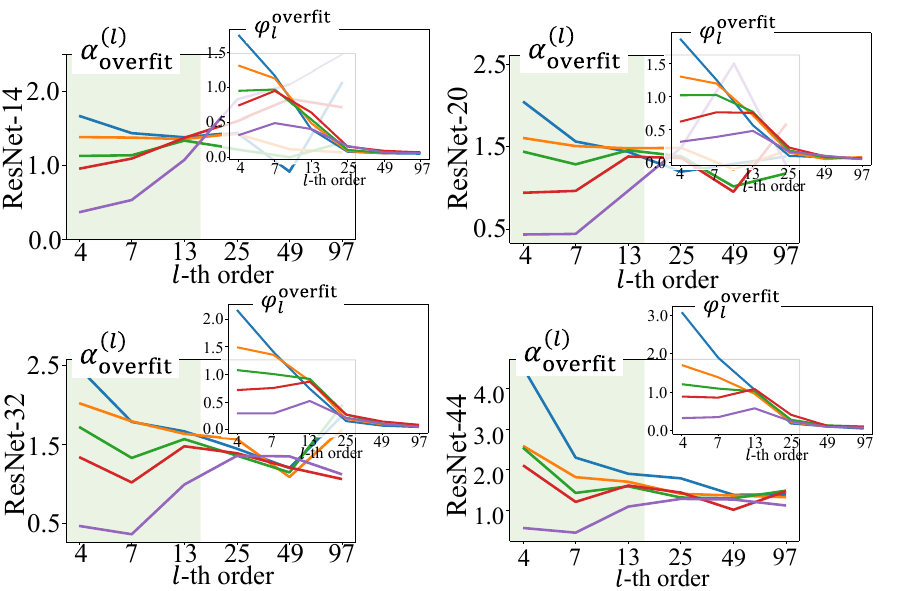}}
		\vspace{-5pt}
		\caption{(left) Effectiveness of feature components {\small$\alpha^{(l)}_{\mathrm{effective}}$}. The top-right sub-figure shows the Shapley value {\small$\varphi_l^\textrm{train}$}; (right) Confidence of feature components being over-fitted {\small$\alpha^{(l)}_{\mathrm{overfit}}$}. The top-right sub-figure shows the Shapley value {\small$\varphi_l^\textrm{overfit}$}.}
		\label{fig:analysis}
	\end{small}
\end{figure}

\textbf{Exp. 3, analysis of the effectiveness and the significance of over-fitting of feature components.}
Figure~\ref{fig:analysis} compares the effectiveness $\alpha^{(l)}_{\textrm{effective}}$ and the significance of over-fitting $\alpha^{(l)}_{\textrm{overfit}}$ of feature components disentangled in Exp. 2.
We found that \textbf{(1)} \textit{in shallow DNNs (like ResNet-8 and ResNet-14), simple feature components were much more effective than complex feature components. However, in deep DNNs, feature components of medium complexity orders tended to be the most effective.} This indicated that the effectiveness of feature components was determined by the network architecture. 
\textbf{(2)} \textit{Simple feature components learned from a small number of samples were usually more over-fitted than simple feature components learned from many samples.} \textbf{(3)} \textit{There was no clear regulation for the significance of over-fitting for high-complexity feature components.} This might be due to the low effectiveness of high-complexity feature components.

\begin{figure}[t]
	\begin{small}
	\centering
	\begin{minipage}{0.33\linewidth}
		\centering
		\includegraphics[width=0.9\linewidth]{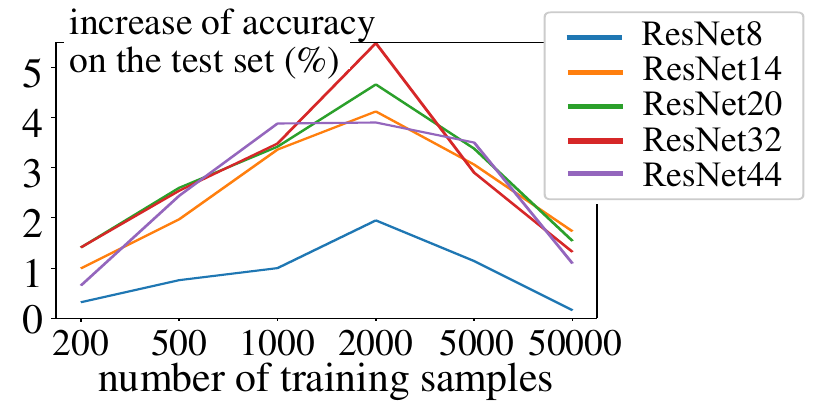}
		\caption{Increase of the classification accuracy based on {\small$\Phi^{(l)}(x)$}.}
		\label{fig:improvement}
		\end{minipage}
		\hfill
	\begin{minipage}{0.6\linewidth}
		\begin{center}
			\captionof{table}{The mean absolute value of prediction errors.}
			\resizebox{\linewidth}{!}{\
						\begin{tabular}{c| c c | c c}
							\hline
							\multirow{3}*{}&  \multicolumn{2}{c|}{Accuracy} & \multicolumn{2}{c}{Task loss}\\
							\cline{2-5}
							{} & Prediction & Range & Prediction & Range \\
							{} & error & of value & error &of value\\
							\hline
							CIFAR-10 & \textbf{2.73\%}&28.73\%-72.83\% & \textbf{0.49} & 1.59-6.42\\
							CUB200 & \textbf{5.66\%}&28.18\%-56.18\% & \textbf{0.47} & 2.94-5.76\\
							Dogs & \textbf{3.26\%}&9.37\%-37.95\% & \textbf{0.34} & 4.34-7.97\\
							\hline
						\end{tabular}}
						\label{tab:regression_loss_acc}
				\end{center}
			\end{minipage}
		\end{small}
\end{figure}

\textbf{Exp. 4, improvement of the classification accuracy based on $\Phi^{(l)}(x)$.}
We further tested the  classification accuracy of ResNets learned on the CIFAR-10 dataset by directly putting $\Phi^{(l)}(x)$ (here $l=7$) into the trained ResNet to replace the original feature $f(x)$.
Figure~\ref{fig:improvement} shows that $\Phi^{(l)}(x)$ further increased the classification accuracy.

\begin{figure}[t]
	\begin{small}
	\begin{minipage}{0.6\linewidth}
		\centering
		\includegraphics[height=0.35\linewidth]{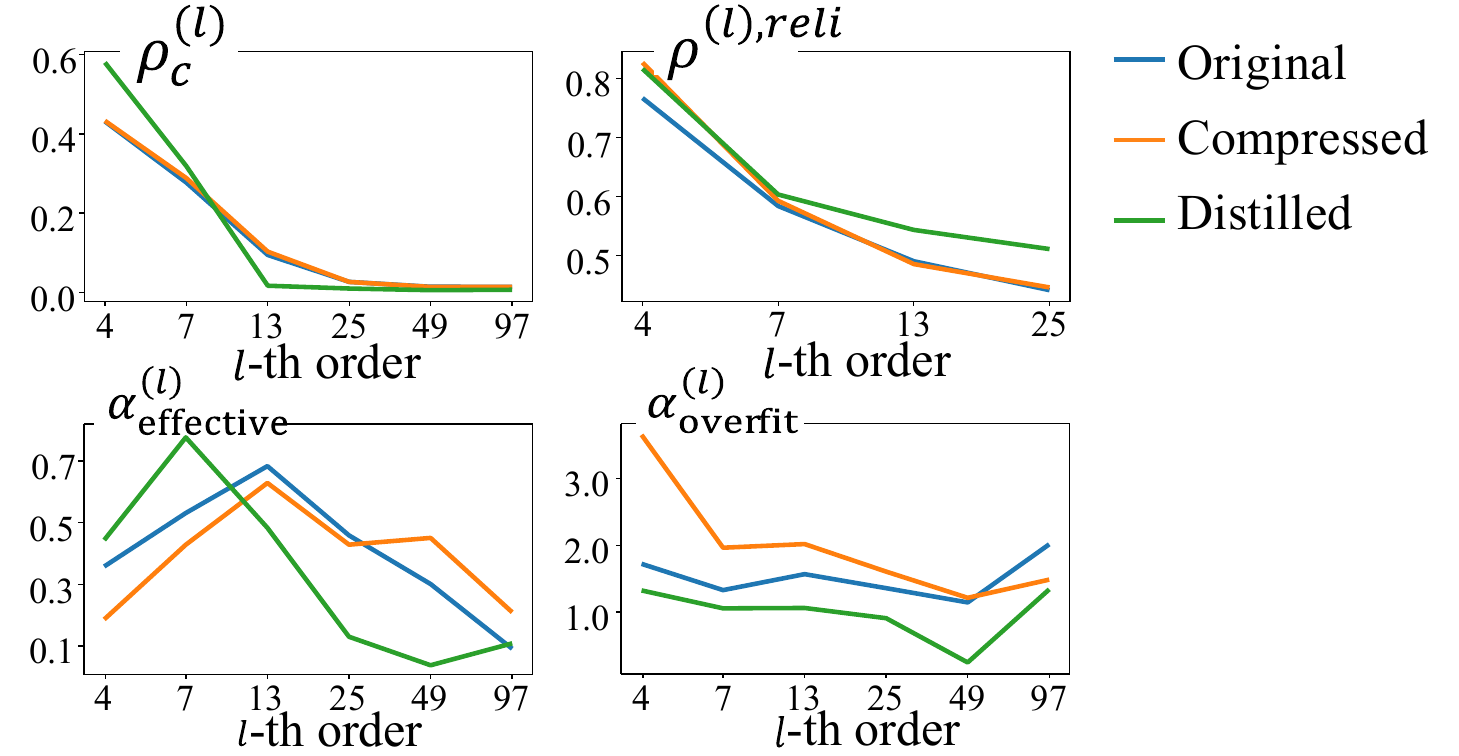}
		\caption{Comparisons of feature complexity between the original DNN, the compressed DNN and the distilled DNN.}
		\vspace{-5pt}
		\label{fig:comparison}
	\end{minipage}
	\hfill
	\begin{minipage}{.37\linewidth}
		\centering
		\vspace{18pt}
		\includegraphics[height=0.45\linewidth]{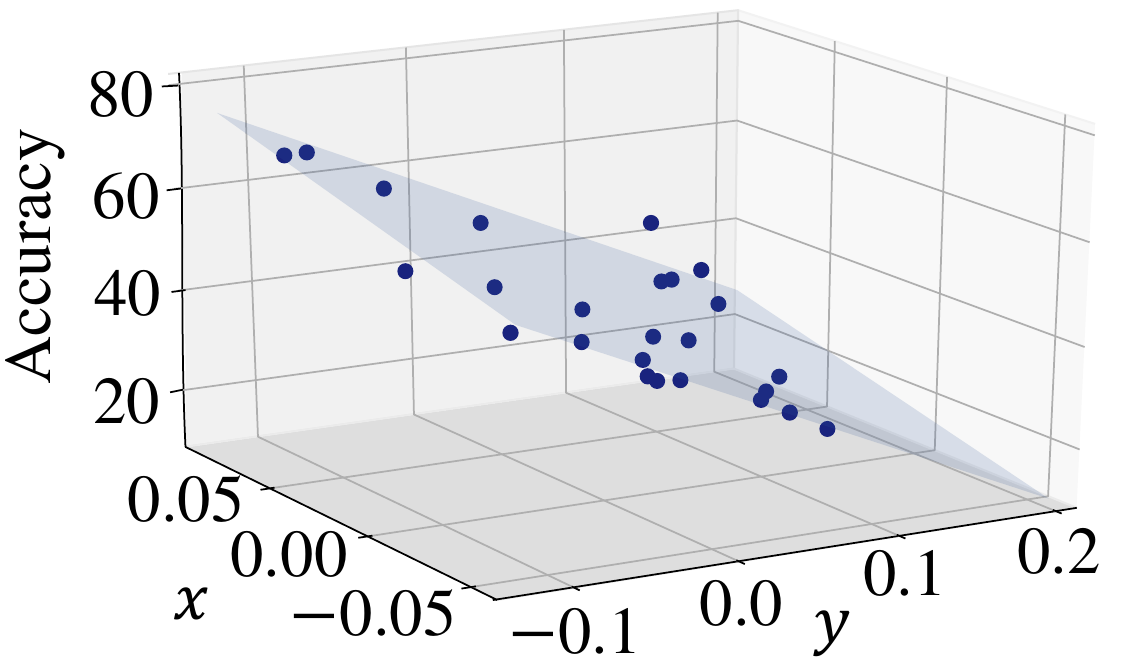}
		\caption{Relationship between the feature complexity and the accuracy.}
		\vspace{-5pt}
		\label{fig:regression}
	\end{minipage}
	\end{small}
\end{figure}

\textbf{Exp. 5, analysis of network compression and knowledge distillation.}
We learned the ResNet-32 on the CIFAR-10 dataset as the originial DNN. We  used the compression algorithm~\cite{han2015deep} to learn another DNN (termed the \textit{compressed DNN}) by pruning and quantizing the trained original DNN. For the knowledge distillation, we used another network (termed the \textit{distilled DNN})\footnote{The distilled DNN had the same architecture with the disentangler net with 7 ReLU layers.}, to distill~\cite{hinton2015distilling} the output feature of the last residual block in the original DNN. The supplementary material provides more details about the network compression and knowledge distillaion in \cite{han2015deep,hinton2015distilling}. We compared the compressed DNN and the distilled DNN with the original DNN.
We disentangled feature components from the output feature of the last residual block in the original DNN and the compressed DNN, and the output feature of the distilled DNN.
Figure~\ref{fig:comparison} shows $\rho_c^{(l)}, \rho^{(l),\textrm{reli}}, \alpha^{(l)}_\textrm{effective},$ and $\alpha^{(l)}_\textrm{overfit}$ in the three DNNs.
For the compressed DNN, \textbf{(1)} \textit{the network compression did not affect the distribution of feature components and their reliability.} \textbf{(2)} \textit{Simple feature components in the compressed DNN exhibited lower effectiveness and higher significance of over-fitting than simple feature components in the original DNN.}
For the knowledge distillation, \textbf{(1)} \textit{the distilled DNN had more feature components of low complexity orders than the original DNN. The simple feature components in the distilled DNN were more effective than those in the original DNN.} \textbf{(2)} \textit{Complex feature components in the distilled DNN were more reliable and less over-fitted than complex feature components in the original DNN.}
These results demonstrated that the knowledge distillation would help DNNs learn more reliable features, which prevented over-fitting.

\textbf{Exp. 6, strong connections between feature complexity and performance of DNNs.}
To this end, we learned a regression model,  which used the distribution of feature components of different complexity orders to predict the performance of DNNs.
For each DNN, we used disentangler nets with $l=4,7,13,25$ to disentangle out {\small$\Phi^{(l),\textrm{reli}}(x)$} and {\small$\Phi^{(l),\textrm{unreli}}(x)$}. Then, we calculated {\small$Var[\Phi^{(l),\textrm{reli}}(x)\!-\!\Phi^{(l-1),\textrm{reli}}(x)]/Var[f(x)]$} and {\small$Var[\Phi^{(l),\textrm{unreli}}(x)\!\!-\!\!\Phi^{(l-1),\textrm{unreli}}(x)]/Var[f(x)]$} for $l=4,7,13,25$, thereby obtaining an 8-dimensional feature to represent the distribution of different feature components. In this way, we learned a linear regressor to use the 8-dimensional feature to predict the testing loss or the classification accuracy. 
For the CIFAR-10 dataset, we applied cross validation: we randomly selected 20 DNNs from 25 pre-trained ResNet-8/14/20/32/44 models on different training sets in Exp. 2 to learn the regressor and used the other 5 DNNs for testing.\footnote{For the CUB200-2011 dataset and the Stanford Dogs dataset, we randomly selected 11 models from 12 pre-trained ResNet-18/34 and VGG-16 models to learn the regressor. One model was used for testing.}
These 25 DNNs were learned using 200-5000 samples, which were randomly sampled from the CIFAR-10 dataset to boost the model diversity.
We repeated such experiments for 1000 times for cross validation.

Table~\ref{tab:regression_loss_acc} reports the mean absolute value of prediction error for the classification accuracy and the task loss  over 1000 repeated experiments. \textit{Linear weights for reliable and unreliable components are shown in supplementary materials.}  The prediction error was much less than the value gap of the testing accuracy and the value gap of the task loss, which indicated the strong connection between the distribution of feature complexity and the performance of DNNs.
 
Figure~\ref{fig:regression} further visualizes the plane of the linear regressor learned on the CIFAR-10 dataset. The visualization was conducted by using PCA~\cite{wold1987principal} to reduce the 8-dimensional feature into a 2-dimensional space, \emph{i.e.} $(x,y)$ in Figure~\ref{fig:regression}. There was a close relationship between the distribution of feature complexity and the performance of a DNN. Please see the supplementary material for more details.

\section{Conclusion}
In this paper, we have proposed a generic definition of the feature complexity of DNNs. We design a method to disentangle and quantify feature components of different complexity orders, and analyze the disentangled feature components from three perspectives. Then, a close relationship between the feature complexity and the performance of DNNs is discovered. Furthermore, the disentangled feature components can improve the classification accuracy of DNNs. As a generic mathematical tool, the feature complexity provides a new perspective to explain existing deep-learning techniques, which has been validated by experiments.

\bibliography{example_paper}
\bibliographystyle{plainnat}

\newpage

\appendix
\section{Theoretical foundations of Shapley values}

This section introduces theoretical foundations of Shapley values mentioned in Section 3.1. The Shapley value was originally proposed in the game theory~\cite{shapley1953value}. Let us consider a game with multiple players. Each player can participate in the game and receive a reward individually. Besides, some players can form a coalition and play together to pursue a higher reward. Different players in a coalition usually contribute differently to the game, thereby being assigned with different proportions of the coalition's reward. The Shapley value is considered as a unique method that fairly allocates the reward to players with certain desirable properties~\cite{ancona2019explaining}. Let $N=\{1,2,\dots,n\}$ denote the set of all players, and $2^{N}$ represents all potential subsets of $N$. A game $v:2^N \to \mathbb{R}$ is implemented as a function that mapping from a subset to a real number. When a subset of players $S\subseteq N$ plays the game, the subset can obtain a reward $v(S)$. Specifically, $v(\emptyset)=0$. The Shapley value of the $i$-th player $\phi^N_{i,v}$ can be considered as an unbiased contribution of the $i$-th player.

\[
\phi^N_{i,v}=\sum_{S\subseteq N\backslash\{i\}}
\frac{(n-|S|-1)!|S|!}{n!}\bigg[v(S\cup\{i\})-v(S)\bigg]
\]

Weber \emph{et al.}~\cite{weber1988probabilistic} have proved that the Shapley value is the only reward with the following axioms.

\textbf{Linearity axiom:} If the reward of a game $u$ satisfies $u(S)=v(S)+w(S)$, where $v$ and $w$ are another two games. Then the Shapley value of each player $i\in N$ in the game $u$ is the sum of Shapley values of the player $i$ in the game $v$ and $w$, \emph{i.e.} $\phi^N_{i,u}=\phi^N_{i,v}+\phi^N_{i,w}$.

\textbf{Dummy axiom:} The dummy player is defined as the player that satisfies $\forall S\subseteq N\backslash\{i\}$, $v(S\cup\{i\})=v(S)+v(\{i\})$. In this way, the dummy player $i$ satisfies $v(\{i\})=\phi^N_{i,v}$, \emph{i.e.} the dummy player has no interaction with other players in $N$.

\textbf{Symmetry axiom:} If $\forall S\subseteq N\backslash\{i,j\}$, $v(S\cup\{i\})=v(S\cup\{j\})$, then $\phi^N_{i,v}=\phi^N_{j,v}$.

\textbf{Efficiency axiom:} $\sum\limits_{i\in N}\phi^N_{i,v}=v(N)$. The efficiency axiom can ensure the overall reward can be distributed to each player in the game.

\section{About the  reliable of features}

This section explains the rationality and implementation details of the algorithm in Section 3.2. For each DNN, we select the feature of a specific intermediate layer as the target feature. Let $f_1(x), f_2(x), \dots, f_K(x)$ denote target features of $K$ DNNs. We aim to extract reliable features of different complexity orders in the $K$ DNNs, \emph{i.e.} $\Phi_1^{(l),\text{reli}}(x), \Phi_2^{(l),\text{reli}}(x),\dots,\Phi_K^{(l),\text{reli}}(x)$. 

Inspired by \cite{liang2019knowledge}, we consider each pair of the $K$ reliable feature components are able to reconstruct each other by a linear transformation. As is mentioned in Section 3.2, $\psi^{(l)}(x)$ is the output of a disentangler net with $l$ ReLU layers. We add $K$ parallel additional convolutional layers $g_1,g_2,\dots,g_K$ on top of $\psi^{(l)}(x)$ to mimic $f_1(x), f_2(x), \dots, f_K(x)$. At this time, their outputs $\widetilde{\Phi}^{(l)}_k (x)=g_k(\psi^{(l)}(x))$ are not able to reconstruct each other linearly.

To enable $\widetilde{\Phi}^{(l)}_i (x)$ and $\widetilde{\Phi}^{(l)}_j (x)$ to reconstruct each other linearly, we first transform $\widetilde{\Phi}^{(l)}_i (x)$ to $\psi^{(l)}(x)$ by the linear regressor $h_i$, and then use $\psi^{(l)}(x)$ to reconstruct $\widetilde{\Phi}^{(l)}_j (x)$ by another linear regressor $g_j$, as shown in Figure~\ref{fig:reli} in the supplementary material. Similarly, we can use $\widetilde{\Phi}^{(l)}_j (x)$ to regress $\psi^{(l)}(x)$ via $h_j$, and then use $\psi^{(l)}(x)$ to linearly regress $\widetilde{\Phi}^{(l)}_i (x)$.

\begin{figure}[h]
    \centering
    \includegraphics[width=0.5\linewidth]{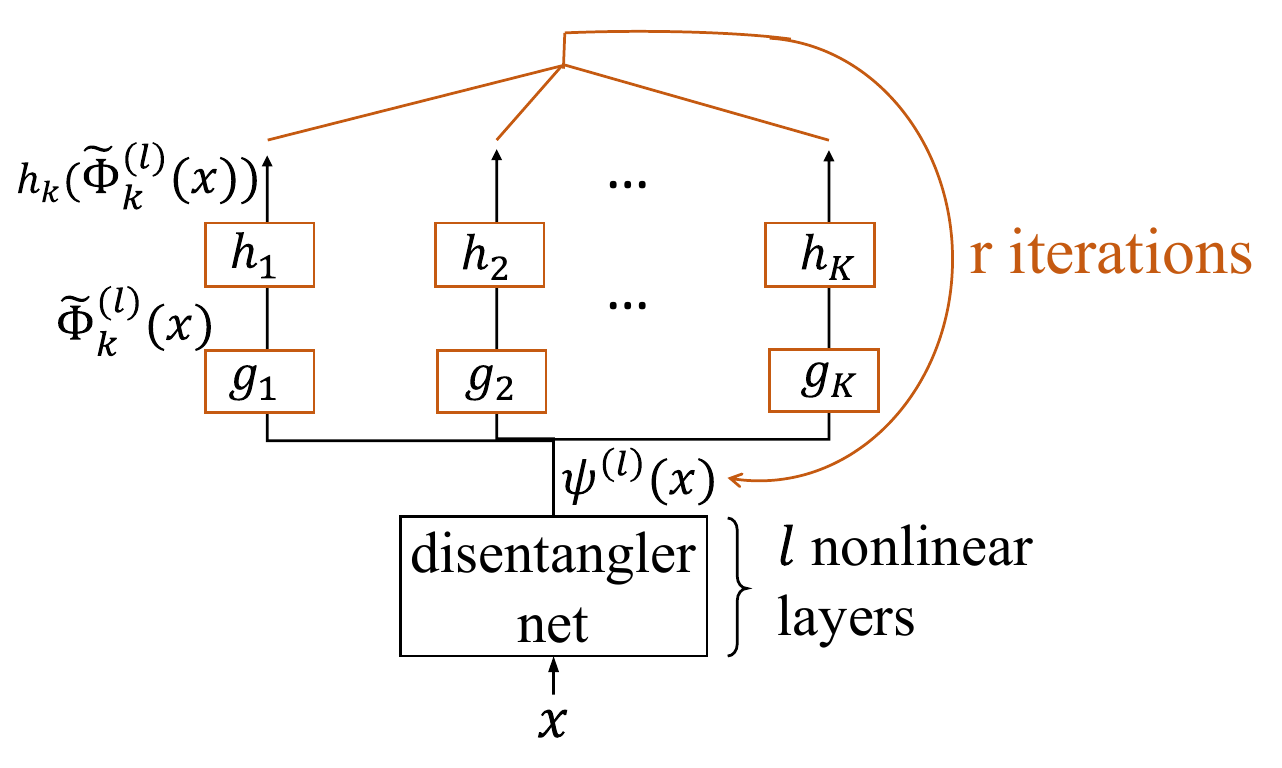}
    \caption{The network for the disentanglement
of reliable feature components.}
    \label{fig:reli}
    \vspace{-10pt}
\end{figure}

In this way, $\tilde{\Phi}^{(l)}_i(x)$ and $\tilde{\Phi}^{(l)}_j(x)$ can reconstruct each other by a linear transformation:
\[
    \tilde{\Phi}^{(l)}_i(x)=g_i ( h_j ( \tilde{\Phi}^{(l)}_j(x))),\quad \tilde{\Phi}^{(l)}_j(x)=g_j ( h_i ( \tilde{\Phi}^{(l)}_i(x)))
\]
We repeat the reconstruction for $R$ iterations and design the loss in Eq. (5).

In the implementation, we first train $g_1,g_2,\dots,g_K$, and then fix $g$ to train $h_1,h_2,\dots,h_K$. To reduce the computational complexity, we did not explicitly optimize on the loss in Eq. (5) which requires a sum over $r$, from 1 to $R$. Instead, in each $r$-th training phase, we optimize $\sum_{k=1}^{K}\Vert h_k\circ g_k \circ \psi^{(l)}_r(x)-\psi^{(l)}_r(x)\Vert^2, r\in \{1,2,\dots, R\}$.

\section{Accuracy of DNNs}

This section contains more details of DNNs in Exp. 2. We trained ResNet-8/14/20/32/44 based on the CIFAR-10 dataset~\cite{krizhevsky2009learning}, and trained VGG-16, ResNet-18/34 based on the CUB200-2011 dataset~\cite{wah2011caltech} and the Stanford Dogs dataset~\cite{KhoslaYaoJayadevaprakashFeiFei_FGVC2011}. More specifically, we trained each DNN with different numbers of training samples, which were randomly sampled from the training set. All teacher networks were pre-trained with different parameter initializations. Table~\ref{tab: acc_loss} in the supplementary material reports the accuracy and loss of the prediction on the testing samples.

\begin{table}[h]
	\caption{Accuracy of DNNs on different datasets.}
	\begin{minipage}{\linewidth}
		\vspace{5pt}
		\centering {\small (a) On the CIFAR-10 dataset.}\\
		\resizebox{0.7\linewidth}{!}{\
			\begin{tabular}{c| c c c c c}
				\hline
				{} & \multicolumn{5}{c}{Accuracy}\\
				\cline{2-6}
				\# of training samples & 200 & 500 & 1000 & 2000 & 5000\\ \hline
				ResNet-8 & 31.37\% & 39.55\% & 45.08\% & 53.82\% & 67.80\%
				\\
				ResNet-14 & 31.50\% & 39.21\% & 47.71\% & 52.41\% & 68.30\% 
				\\
				ResNet-20 & 31.56\% & 38.40\% & 46.09\% & 56.15\% & 70.62\% 
				\\
				ResNet-32 & 30.48\% & 37.94\% & 46.71\% & 56.80\% & 72.83\% 
				\\
				ResNet-44 & 28.73\% & 38.57\% & 46.63\% & 56.00\% & 70.66\%
				\\
				\hline				
			\end{tabular}}
	\end{minipage}	
	\begin{minipage}{\linewidth}
		\vspace{5pt}
		\centering {\small (b) On the CUB200-2011 dataset.}\\
		\resizebox{.6\linewidth}{!}{\
			\begin{tabular}{c| c c c c }
			\hline
			{} & \multicolumn{4}{c}{Accuracy}\\
			\cline{2-5}
			\# of training samples & 2000 & 3000 & 4000 & 5000\\ \hline
			ResNet-18 & 32.36\% & 44.82\% & 52.73\% & 56.18\% 
			\\
			ResNet-34 & 29.43\% &43.86\% &	52.17\% & 53.68\%
			\\
			VGG-16 & 28.18\% &41.04\% &	47.03\% & 53.83\%
			\\
			\hline			
		\end{tabular}}
	\end{minipage}
	\begin{minipage}{\linewidth}
		\vspace{5pt}
		\centering {\small (c) On the Stanford Dogs dataset.}\\
		\resizebox{.6\linewidth}{!}{\
		\begin{tabular}{c| c c c c}
			\hline
			{} & \multicolumn{4}{c}{Accuracy} \\
			\cline{2-5}
			\# of trainingsamples & 1200 & 2400 & 3600 & 4800\\ \hline
			ResNet-18 & 10.93\% & 19.42\% & 28.51\% &37.95\% 
			\\
			ResNet-34 & 9.37\% & 18.83\% & 27.05\% & 32.23\%
			\\
			VGG-16 & 10.36\% & 16.78\% &23.63\% &29.14\%
			\\
			\hline			
		\end{tabular}}
	\end{minipage}
	\label{tab: acc_loss}
\end{table}

\section{Architecture of the disentangler net}

\begin{small}
	\begin{figure}[h]
		\centering
		\includegraphics[width=0.6\linewidth]{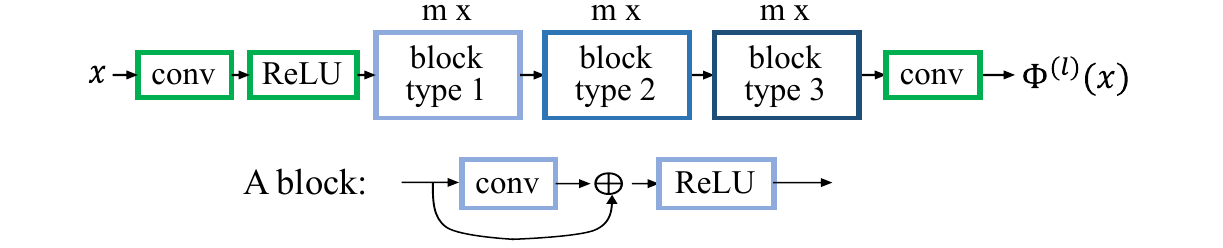}
		\caption{Disentangler net.}
		\label{fig:disentangler networks}
		\end{figure}
\end{small}

\section{Visualization of feature components}

This section shows more visualization results in Section 4 by visualizing feature components disentangled from the target feature. For better visualization, we took the output feature of \texttt{conv4-3} (with shape $28\times28\times512$) in VGG-16 as the target feature. In the following figures, we considered the most activated channel in the target feature map as $f(x)$. Then, we disentangled and visualized $f(x)$, $c^{(l)}(x)$ and $\Phi^{(l)}(x)$. We found that  
low-complexity feature components usually represented the general shape of objects, while complex feature components
corresponded to detailed shape and noises.

	\begin{figure}[h]
		\centering
		\includegraphics[width=0.8\linewidth]{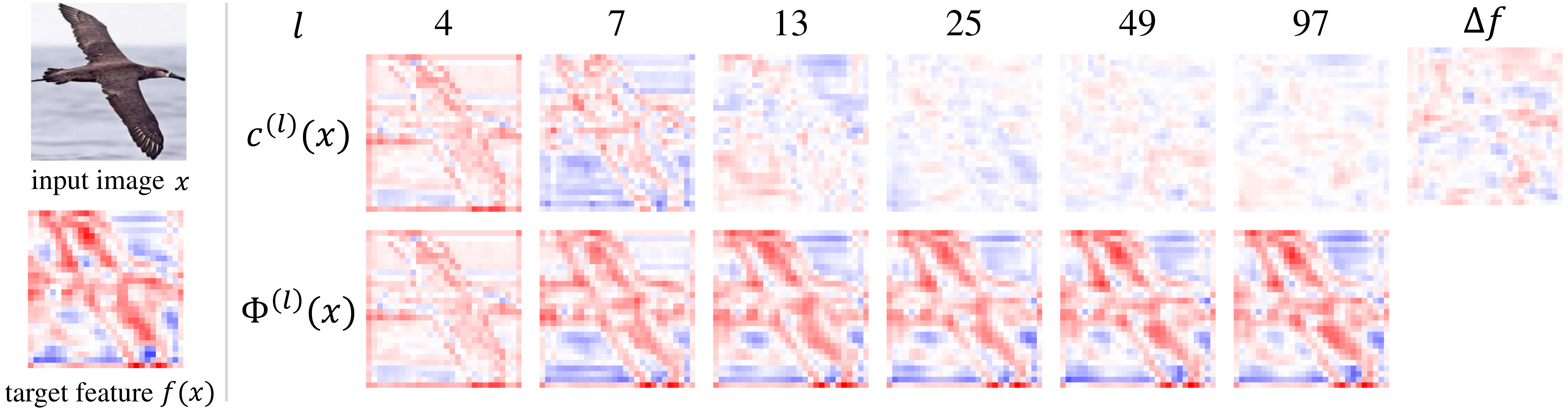}
	\end{figure}
	\begin{figure}[h]
		\centering
		\includegraphics[width=0.8\linewidth]{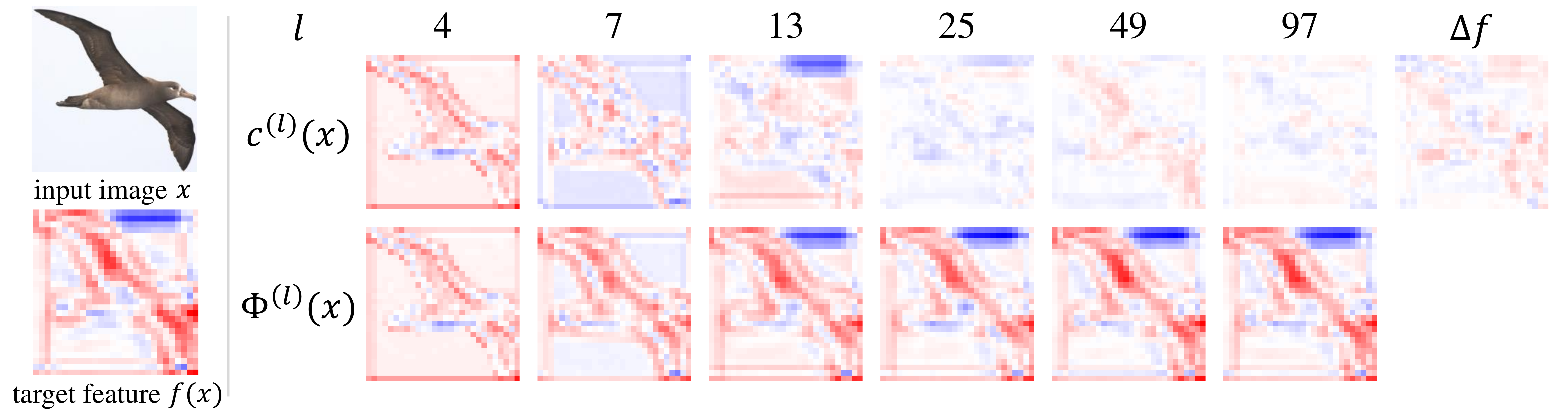}
	\end{figure}
	\begin{figure}[h]
		\centering
		\includegraphics[width=0.8\linewidth]{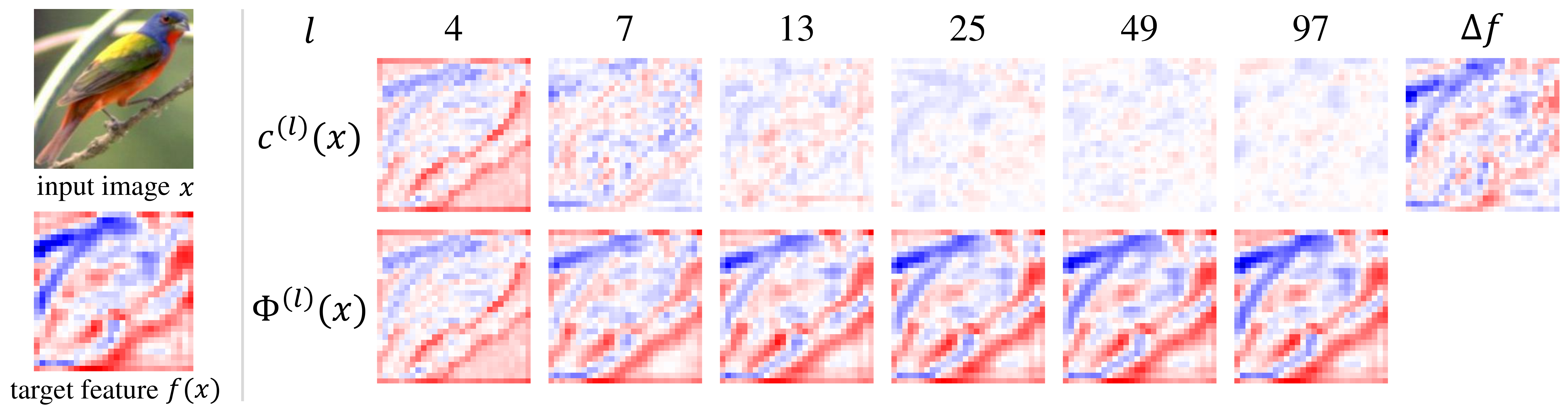}
	\end{figure}
	\begin{figure}[h]
		\centering
		\includegraphics[width=0.8\linewidth]{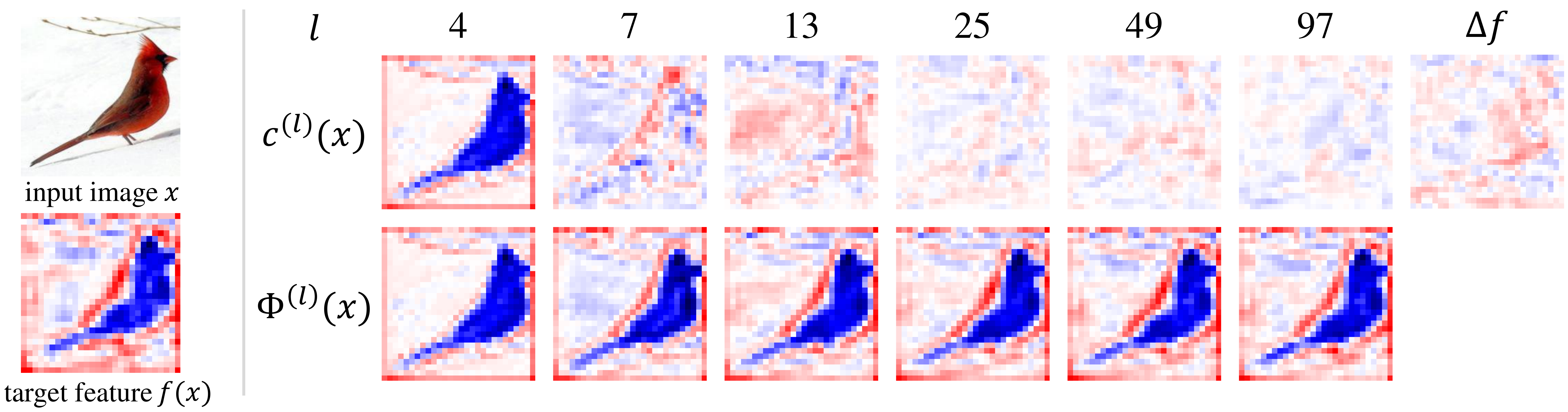}
	\end{figure}
	\begin{figure}[h]
		\centering
		\includegraphics[width=0.8\linewidth]{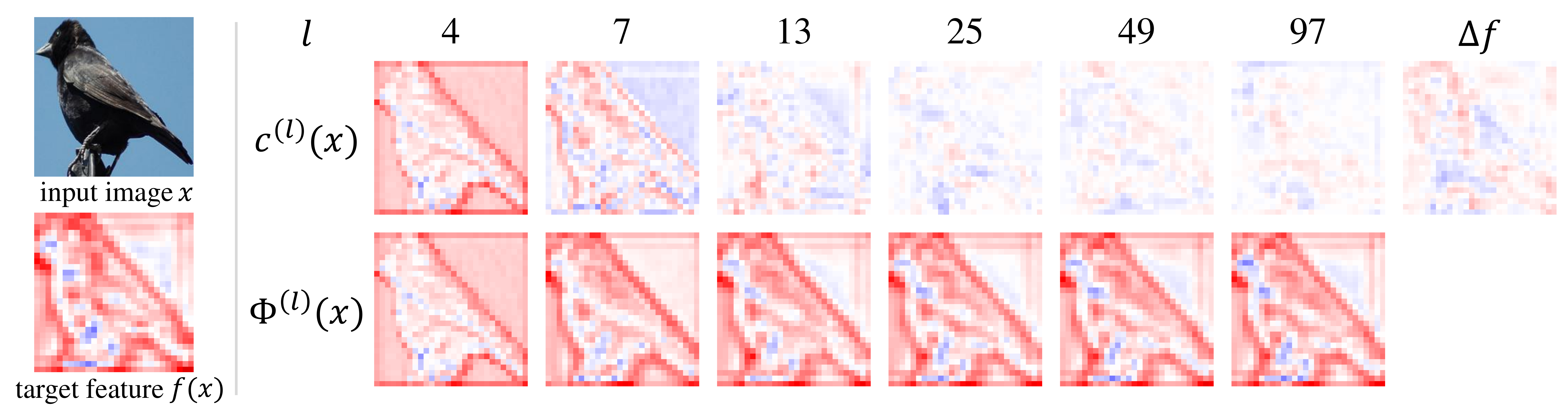}
	\end{figure}
	\begin{figure}[h]
		\centering
		\includegraphics[width=0.8\linewidth]{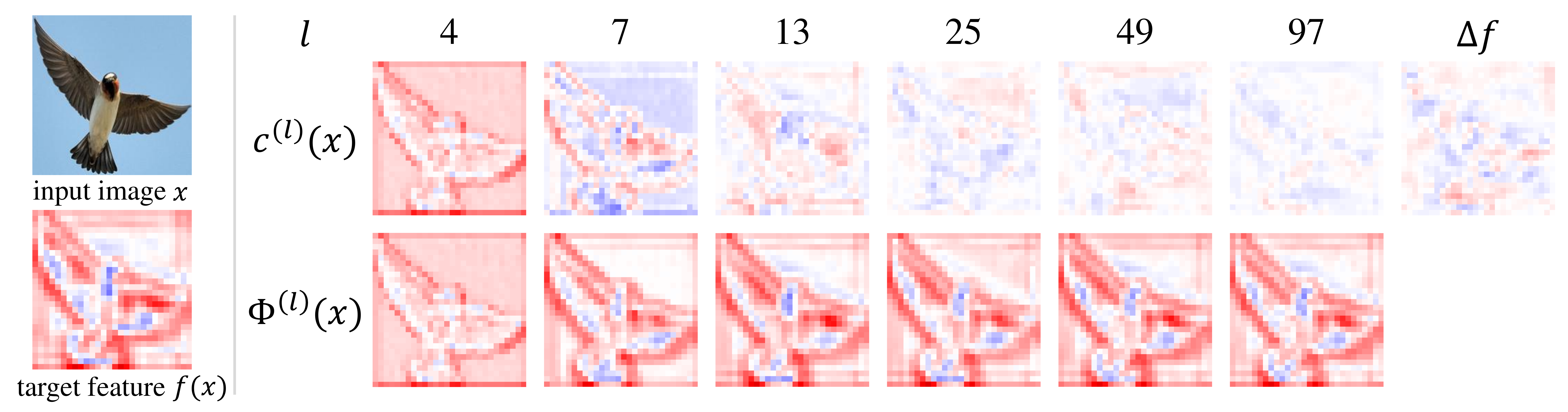}
	\end{figure}
	\begin{figure}[h]
		\centering
		\includegraphics[width=0.8\linewidth]{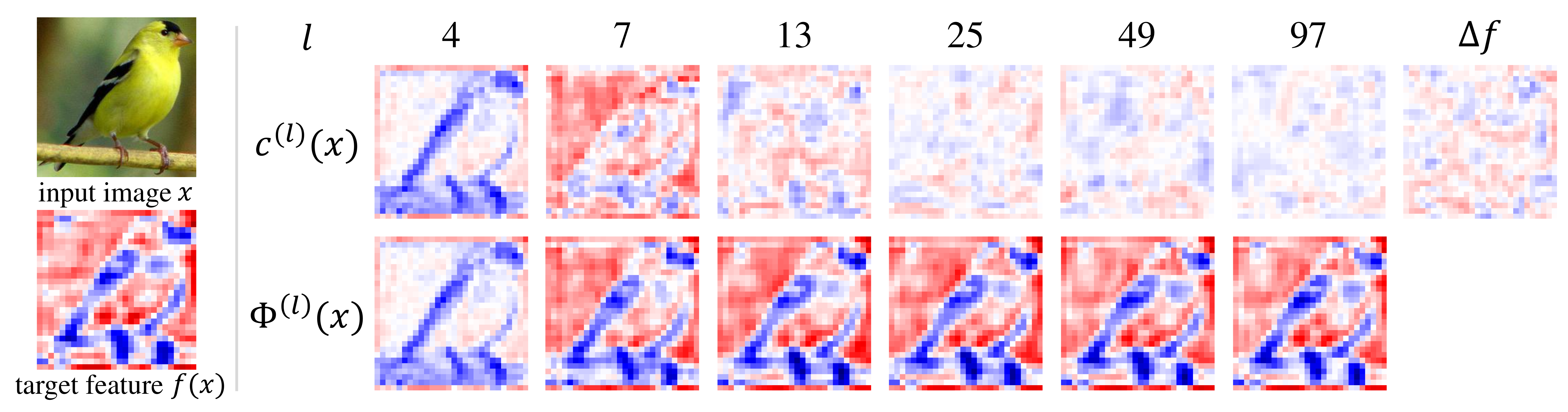}
	\end{figure}
	
	\clearpage
	
\section{Evaluation of DNNs learned on the CUB200-2011 dataset and the Stanford Dogs dataset.}

This section shows more experimental results on different datasets in Exp. 2. Figure \ref{fig:cubdogs} in the supplementary material shows the result of $\rho^{(l)}_c$ of ResNet-18/34 learned on the CUB200-2011 dataset and the Stanford Dogs dataset. We found that the DNN learned from the larger training set usually encoded more complex features, but the overall distribution of feature components was very close to the DNN learned from the smaller training set. This indicated that the number of training samples had small impacts on the significance of feature components of different complexity orders.

\begin{figure}[htbp]
    \centering
    \subfigure[\label{fig:cub}Significance of feature components $\rho^{(l)}_c$ in DNNs learned on CUB200-2011]{\includegraphics[width=.8\linewidth]{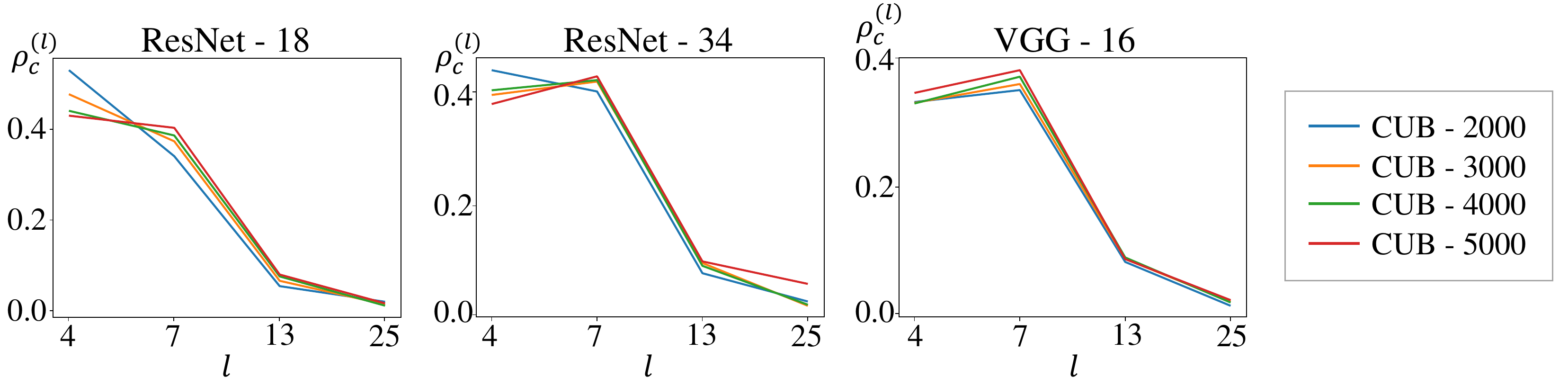}}
    \subfigure[\label{fig:dogs}Significance of feature components $\rho^{(l)}_c$ in DNNs learned on Stanford Dogs]{\includegraphics[width=.8\linewidth]{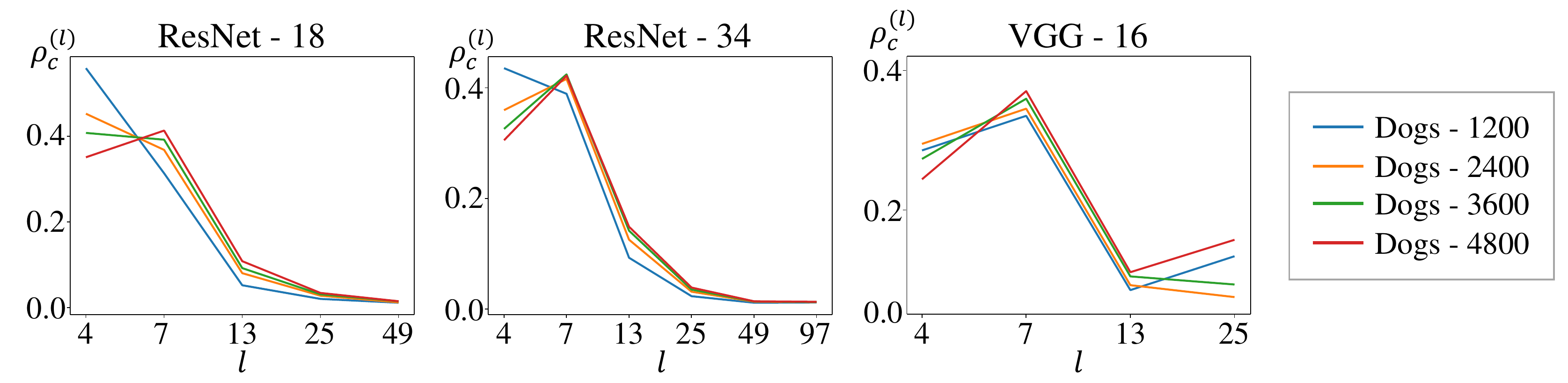}}
    \caption{Significance of feature components in DNNs learned on the CUB200-2011 dataset and the Stanford Dogs dataset.}
    \label{fig:cubdogs}
\end{figure}

\section{Details of network compression and knowledge distillation}

This section introduces more details about the network compression and knowledge distillation in Exp. 5. 

Network compression based on~\cite{han2015deep}: We learned a compressed DNN by pruning and then quantization. In the pruning phase, we pruned the DNN with sensitivity rate 1.0 for all convolutional and fully connected layers. We iteratively pruned the DNN and retrain the weights. The number of this iteration was 300 and the weights were retrained for 20 epochs in our experiments. The weights in the pruned DNN was retrained for 100 epochs. For example, as a result ResNet-32 trained on CIFAR10-1000 had an overall pruning rate of 5.88$\times$ without affecting the accuracy significantly. To compress further, we quantized weights in the DNN. For weights in convolutional layers, we quantized them to 8 bits, while for weights in fully connected layers, we quantized them to 5 bits. In this way, we obtained the compressed DNN.

Knowledge distillation based on~\cite{hinton2015distilling}: To obtain the distilled DNN, we used a shallower DNN to mimic the intermediate-layer feature of the original DNN. For simplicity, we let the distilled DNN have the same architecture with the disentangler net with seven ReLU layers. The distilled DNN usually addressed the problem of over-fitting. For example, the distilled DNN based on ResNet-32 on CIFAR10-1000 had a 3.48\% decrease in testing accuracy, without affecting the training accuracy significantly.

\section{Connection between complexity and performance}
This section introduces experimental details of Exp. 5.
In Exp. 5, we learned a regression model  to predict the performance of the DNN with the distribution of feature components of different complexity orders.
For each DNN, we used disentangler networks with $l=4,7,13,25$ to disentangle out {\small$\Phi^{(l),\textrm{reli}}(x)$} and {\small$\Phi^{(l),\textrm{unreli}}(x)$}. Then, we calculated {\small$Var[\Phi^{(l),\textrm{reli}}(x)\!-\!\Phi^{(l-1),\textrm{reli}}(x)]/Var[f(x)]$} and {\small$Var[\Phi^{(l),\textrm{unreli}}(x)\!\!-\!\!\Phi^{(l-1),\textrm{unreli}}(x)]/Var[f(x)]$} for $l=4,7,13,25$, thereby obtaining an 8-dimensional feature to represent the distribution of different feature components. In this way, we learned a linear regressor to use the 8-d feature to predict the testing loss or the classification accuracy, as follows.
\begin{small}
\[
\textrm{result} = \sum_{i=1}^{4} \alpha_i \times \frac{Var[\Phi^{(l_i), \textrm{reli}}(x)-\Phi^{(l_{(i-1)}), \textrm{reli}}(x)]}{Var[f(x)]} + \sum_{i=1}^{4} \beta_i \times \frac{Var[\Phi^{(l_i), \textrm{unreli}}(x)-\Phi^{(l_{(i-1)}), \textrm{unreli}}(x)]}{Var[f(x)]} + b
\]
\end{small}
where $l_i\in \{4,7,13,25\}$.

\textbf{Testing on the compressed DNNs:} Just like in Exp. 5, we also conducted experiments to show the strong connection between the feature complexity and the network performance in terms of the compressed DNNs.
We predicted the accuracy of the compressed DNNs with the learned regressor and Table~\ref{tab:compressionresult} in the supplementary material shows the result. The prediction error was relatively small, and it further validated the close relationship between the feature complexity and the performance of DNNs.

\begin{table}[h]
	\caption{Prediction result of the accuracy of the compressed DNNs.}
	\resizebox{\textwidth}{!}{%
		\begin{tabular}{c|c c c c|c c c}
			\hline
			Model                                           & ResNet-14 & ResNet-20 & Resnet-32 & ResNet-44 & \multicolumn{3}{c}{ResNet-32}         \\ \hline
			Dataset                                         & \multicolumn{4}{c|}{CIFAR10-2000}              & CIFAR10-500 & CIFAR10-1000 & CIFAR10-5000 \\ \hline
			Test acc. before compression          & 52.41     & 56.15     & 56.80     & 56.00     & 37.94      & 46.71       & 72.83       \\
			Test acc. after compression           & 53.88     & 57.94     & 60.81     & 58.04     & 38.35      & 49.86       & 73.62       \\
			Predicted acc. after compression & 50.97     & 54.32     & 60.46     & 57.41     & 45.23      & 53.35       & 72.85       \\
			Error                                           & -2.91     & -3.62     & -0.35     & -0.63     & +6.88      & +3.49       & -0.77       \\ \hline
		\end{tabular}%
	}
	\label{tab:compressionresult}
\end{table}

\end{document}